\ifdefined\pdfoutput\pdfoutput=1\fi

\documentclass[11pt,a4paper]{article}

\usepackage[margin=25mm]{geometry}
\usepackage{amsmath}
\usepackage{amssymb}
\usepackage{comment}
\usepackage{tabularx}
\usepackage{array}
\usepackage{etoolbox}
\usepackage{graphicx}
\usepackage{xcolor}
\usepackage[font=small,labelfont=bf,labelsep=period]{caption}
\usepackage[colorlinks=true,allcolors=blue]{hyperref}

\newcommand{\articletype}[1]{}
\renewcommand{\title}[1]{\begin{center}\LARGE\bfseries #1\end{center}\vspace{1mm}}
\renewcommand{\author}[1]{\begin{center}\large #1\end{center}\vspace{-2mm}}
\newcommand{\affil}[1]{\begin{center}\small #1\end{center}\vspace{-4mm}}
\newcommand{\email}[1]{\begin{center}\small\textbf{E-mail:} #1\end{center}\vspace{2mm}}
\newcommand{\keywords}[1]{\noindent{\small\textbf{Keywords:} #1}\par\vspace{2mm}}
\newcommand{\orcid}[1]{\href{https://orcid.org/#1}{\includegraphics[width=8pt]{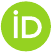}}}

\newcommand{\data}[1]{\section*{Data availability}#1}
\newcommand{\roles}[1]{\section*{Author contributions}#1}

\AtBeginEnvironment{tabular}{\fontsize{8}{10}\selectfont}
\AtBeginEnvironment{tabularx}{\fontsize{8}{10}\selectfont}

\begin{document}

\articletype{Paper}

\title{Reconstructing Backpropagation from Forward Fluctuations in Noise-modulated Neural Networks}

\author{Shuhei Ikemoto$^{1,*}$\orcid{0000-0003-4885-8746}}

\affil{$^1$Graduate School of Life Science and Systems Engineering, Kyushu Institute of Technology, Kitakyushu, Japan}

\affil{$^*$Author to whom any correspondence should be addressed.}

\email{s.ikemoto@ieee.org}


\begin{abstract}

A Noise-modulated Neural Network (NNN) is a neural network that can only learn and infer by adding noise, utilizing the noise as a computational resource rather than eliminating it as a disturbance. 
Thanks to the injected noise, this model can learn efficiently using backpropagation while employing spike-like signals. 
However, backpropagation requires a reverse path via the transposed weight matrices, known as the weight transport problem. 
Consequently, relying on backpropagation weakens its biological or neuromorphic plausibility. 
Many attempts have been made to avoid this weight transport and to train using only the forward pass. 
However, most of them replace the gradients of backpropagation with alternative objective functions or fixed random weights, which can introduce errors in credit assignment. 
In this paper, I demonstrate that the structure of backpropagation in the NNN can be reconstructed solely from forward-pass statistics. 
By using the weight mirror that estimates the weight matrix from the covariance between a previous-layer unit's output and the next-layer unit's input, and combining this with local differential estimation within the units, the output error can be propagated recursively along the computational graph. 
I reconstruct the weight updates of backpropagation without any transposed weight readouts. 
The update of the proposed learning rule closely agrees with that of backpropagation in regression experiments. With Adam updates for each local weight, it achieves final accuracy comparable to backpropagation for simple regression tasks and scales to MNIST classification with up to four hidden layers. 
Furthermore, when the injected noise is uniform, the local operations reduce to the polynomial and comparator levels, and the entire system, including the learning rule, can be expressed with standard digital elements. 
These results demonstrate that in the NNN, noise serves as a resource not only for inference but also for reconstructing backpropagation.

\end{abstract}

\section{Introduction}
\label{sec:intro}

In biological nervous systems, neural noise, such as trial-to-trial variability in spike timing and firing rates, has been considered to bear functional roles rather than merely degrading signals. 
Representative examples include stochastic resonance, where noise helps detect weak signals by crossing a threshold \cite{gammaitoni1998stochastic,moss2004stochastic}, trial-to-trial variability enhancing exploration and learning \cite{fiete2006gradient,fiete2007birdsong}, and stochastic firing serving as the foundation for probabilistic inference and sampling in the brain \cite{pecevski2011probabilistic,orban2016neural}.
These examples share a perspective that treats noise as a computational resource rather than as a nuisance \cite{faisal2008noise,mcdonnell2011benefits}.
Based on this view, establishing a mathematical model that essentially and fully utilizes noise as a computational resource is a key target for neuromorphic computing.

The Noise-modulated Neural Network (NNN) \cite{ikemoto2018noise,ikemoto2021noise,ikemoto2026spatial} brings this perspective into artificial neural networks. 
Its basic elements are stochastic binary units called crossing activation functions, which fire only when the input crosses a threshold. 
The model generates no spikes under constant, noise-free inputs. 
Information transmission via spike-like binary signals occurs only when noise is injected into the units. 
The continuous information of the input vector is probabilistically encoded into these firing events. 
Functions equivalent to those of a conventional neural network (NN) are achieved by decoding the probabilistic features of the output vector through ensemble averaging. 
Furthermore, both the expected output and the expected derivative of the crossing activation function can be estimated from forward propagation samples with noise. 
This estimation requires no prior knowledge of the noise probability distribution.
Consequently, despite using spike-like binary signals, the NNN can be efficiently trained via backpropagation using estimated gradients rather than surrogate gradients\cite{neftci2019surrogate}.
Thus, the NNN trains via backpropagation despite using spike-like signals.

Although backpropagation makes the NNN practical, it undermines its biological plausibility and neuromorphic validity.
Backpropagation requires multiplying the output error by the transposed matrix of the forward weights. 
This process, called weight transport, requires a separate backward data path, a transpose-accessible weight memory, and strict synchronization between phases. 
Consequently, backpropagation is difficult to implement in efficient dedicated hardware \cite{neftci2017event,frenkel2021learning}. 
It also lacks biological plausibility because it requires centralized execution and lacks a known neuroscientific basis \cite{Lillicrap2020backpropagation,crick1989recent}.
Various methods have attempted to solve this issue, such as bypassing weight transport, estimating weights, or using local search \cite{Lillicrap2016random,nokland2016direct,kolen1994backpropagation,akrout2019deep,fiete2006gradient,spall1992multivariate}.
However, these methods face challenges, including training instability from inaccurate gradients and reduced model validity due to retrofitted processes. 
This problem stems from the trade-off between engineering utility and biological plausibility, creating a need for a better compromise.

\begin{figure}
 \centering
    \includegraphics[width=1.0\textwidth]{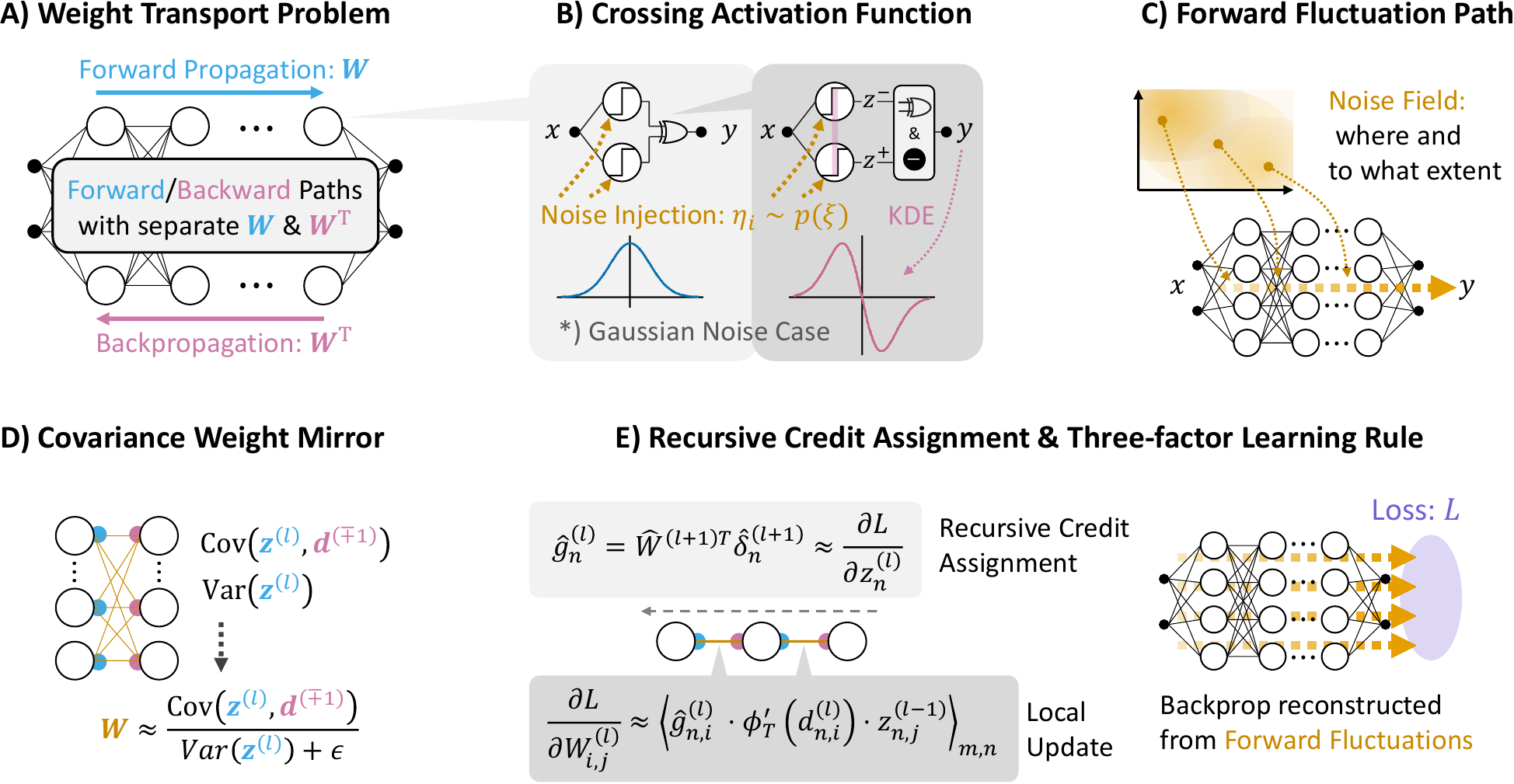}
 \caption{Overview of the problem and proposed approach. (A) Conventional backpropagation requires a backward path and access to transposed forward weights, resulting in the weight transport problem. (B) The crossing activation function generates stochastic binary activity and permits distribution-free estimation of its local derivative by kernel density estimation based on forward noise samples. (C) Noise-induced fluctuations propagate through the network during forward computation. (D) Forward weights are estimated from the covariance between activations and subsequent pre-activations, forming a covariance weight mirror. (E) The estimated weights and local derivatives enable recursive credit assignment and a local three-factor learning rule, thereby reconstructing backpropagation from forward fluctuations.}
\label{fig:overview}
\end{figure}

This study leverages the noise already used in the NNN for learning and inference. 
I show that the structure of backpropagation can be naturally reconstructed solely from forward-pass statistics by exploiting noise-induced fluctuations. 
As summarized in Fig.~\ref{fig:overview}, the proposed approach replaces the transported transposed weights with a covariance-based weight mirror and combines it with local derivative estimation to recursively assign credit.
Specifically, I build on two related mechanisms. The first is the weight mirror, which estimates interlayer weights from the covariance between the activations of the preceding layer and the preactivation variables of the subsequent layer \cite{akrout2019deep}. The second is the Kolen-Pollack method, which maintains alignment between forward and feedback weights through coordinated parameter updates \cite{kolen1994backpropagation}.
By obtaining the required statistics from forward perturbation samples, I integrate the weight mirror without retrofitting dedicated processes. 
This approach allows me to estimate backward weights and derivatives using only forward computation, enabling recursive error propagation along the computation graph. 
Furthermore, I show that this model is suitable for hardware implementation. 
The crossing activation consists of two comparators and an XOR operation regardless of the noise distribution. 
With uniform noise, the noise generator becomes simple, since uniform random numbers can be generated by Xorshift or linear feedback shift registers (LFSRs) with a few XOR and shift operations, without circuits for transformations like Box-Muller. 
Thus, with uniform noise, the entire system, including noise generation and the learning rule, can be expressed with standard digital elements. I quantify the resulting architectural requirements rather than claiming an advantage in power or latency.

In summary, the contributions of this paper are as follows:
\begin{enumerate}
    \item \textbf{A family of covariance credit assignments for NNN:} I systematically construct a learning rule that recursively propagates errors via a covariance weight mirror, derived from analyzing estimator bias.
    \item \textbf{Training accuracy comparable to backpropagation:} I verify that the proposed method provides gradient estimates that closely match backpropagation gradients, achieving equivalent final accuracy in a regression task, and remaining within two percentage points of backpropagation on an MNIST subset with up to four hidden layers.
    \item \textbf{High digital hardware friendliness:} I show that uniform noise reduces activation functions and derivatives to simple expressions, enabling hardware implementation using fewer multipliers, efficient random number generators, and standard digital elements.
\end{enumerate}

The remainder of this paper is organized as follows. 
Section \ref{sec:related_works} reviews related work, and Section \ref{sec:nnn} formalizes the NNN and its activation function. 
Section \ref{sec:forward_only_learning_rule} introduces the forward-only learning rules, followed by experimental validation in Section \ref{sec:experiment}. 
Section~\ref{sec:extended} extends the evaluation. 
Section \ref{sec:hardware_implementation} discusses the hardware suitability of the uniform-noise NNN. 
Section \ref{sec:discussion} provides an overall discussion, and Section \ref{sec:conclusion} concludes the paper.

\section{Related Work}
\label{sec:related_works}

This section reviews three lines of work. The first uses noise as a computational resource. The other two address the weight transport problem of backpropagation in different ways.

\subsection{Neuromorphic Computing with Noise as a Resource}
\label{subsec:2.1}

The idea that noise can be useful goes back to stochastic resonance.
In nonlinear systems, a certain amount of noise enhances the detection and transmission of weak inputs. 
First discovered in climate dynamics \cite{benzi1981mechanism,benzi1982stochastic}, it was later theoretically formalized as a general physical phenomenon \cite{wiesenfeld1995stochastic,gammaitoni1998stochastic,mcdonnell2009stochastic}. 
In biological systems, stochastic resonance is widely demonstrated in sensory pathways such as crayfish mechanoreceptors \cite{douglass1993noise}, cricket cercal systems \cite{levin1996broadband}, and human tactile senses \cite{collins1996noise}.
In the central nervous system, trial-to-trial variability is recognized as a resource for encoding, exploration, and computation rather than mere error \cite{faisal2008noise}. 
In computational neuroscience, stochastic firing is often regarded as sampling for probabilistic inference \cite{hoyer2002interpreting,heafner2016perceptual,berkes2011spontaneous,buesing2011neural}. 
Here, noise serves to generate samples from a posterior distribution. 
In machine learning, noise injection is widely used to stabilize and generalize training \cite{bishop1995training,srivastava2014dropout,neelakantan2015adding}.

Stochastic and spiking neuron models treat noise as a core computational principle rather than an addition to deterministic computation \cite{maass1997networks,neftci2017event}.
They generate discrete spikes probabilistically and convey continuous information through expected values. 
However, their non-differentiable activity is incompatible with gradient-based learning. 
The standard solution is the surrogate gradient method \cite{lee2016training,zenke2021remarkable} or the straight-through estimator \cite{bengio2013estimating}, which replace step-function derivatives with smooth alternatives. 
These surrogate derivatives are chosen by hand and have no link to the forward computation. 
The result is a mismatch between inference and learning, known as the surrogate gradient mismatch \cite{neftci2019surrogate}. 
Alternatively, escape-noise models feature a firing probability that is a smooth function of the membrane potential. 
Thus, gradients of expected firing rates or log-likelihoods can be derived for training \cite{pfister2006optimal,rezende2014stochastic}. 
However, this requires explicitly assuming the analytical form of the firing probability function. 
Randomness is also used in hardware. 
In stochastic computing, multiplication is an AND operation, and addition is multiplexing over bitstreams \cite{alaghi2013survey}.

In these research fields, the NNN is unique because it fully utilizes noise to the extent that it functions only through stochastic resonance, requiring no prior knowledge of the noise distribution \cite{ikemoto2018noise,ikemoto2021noise,ikemoto2026spatial}. 
Furthermore, because both the forward response and its derivative are derived from the same noise statistics, it avoids the surrogate gradient mismatch. 
This study uses this property as a starting point. 
I repurpose the stochastic samples generated for inference to perform credit assignment during learning.

\subsection{Forward Learning Rules with True Gradient Convergence}
\label{subsec:2.2}

Backpropagation requires transposed weight matrices for hidden-layer credit assignment. This weight transport problem has long challenged both biological plausibility and hardware implementation \cite{crick1989recent,Lillicrap2020backpropagation,kunin2020two}.
Approaches to this problem are categorized by whether they recover the true gradient. 
This section reviews methods aiming for true gradient recovery. 
Target propagation \cite{bengio2014how,lee2015difference,ernoult2022towards}, equilibrium propagation \cite{scellier2017equilibrium}, and predictive coding \cite{whittington2017approximation} approximate backpropagation gradients under specific conditions through inverse mappings, equilibrium states, or energy minimization. 
More directly, the Kolen-Pollack method \cite{kolen1994backpropagation} and weight mirrors \cite{akrout2019deep} align weights with the true transpose. 
The Kolen-Pollack method asymptotically matches forward and backward weights via identical updates. 
Weight mirrors directly estimate the transpose using correlations from layer-wise noise injection. 
These methods converge to the true gradient, but they need an extra process that runs outside normal inference and learning. This extra process is what limits their plausibility.
In the NNN, these perturbations are already present in every forward pass. 
I use them to estimate the weights and to propagate errors recursively from forward statistics alone, with no separate process.

\subsection{Simple Forward Learning Rules with Deviations from True Gradients}
\label{subsec:2.3}

The other approach compromises true gradient convergence to seek local and simple update rules that require neither backward data paths nor weight transport. 
Feedback alignment uses a random matrix in place of the transposed weights, and direct feedback alignment projects the output error directly to each hidden layer \cite{Lillicrap2016random,nokland2016direct}. 
The sign-symmetry method, which shares only the signs of forward weights with the backward matrix, also falls into this category \cite{liao2016how,xiao2019biologically}. 
Methods that estimate gradients from correlations between perturbations and loss changes include node perturbation \cite{werfel2005learning,fiete2006gradient}, weight perturbation or SPSA \cite{spall1992multivariate}, REINFORCE-like policy gradients \cite{williams1992simple}, and forward gradient or zeroth-order methods \cite{baydin2022gradients,ren2023scaling}. 

Several methods estimate gradients from forward computations without backpropagation.
Forward-mode automatic differentiation computes the derivative along a random direction and averages such derivatives to estimate the gradient \cite{baydin2022gradients,ren2023scaling}.
Such methods have also been applied to spiking neural networks \cite{wan2026randomized}. 
With Gaussian noise, the covariance between the loss and the perturbation gives, after normalization, a Stein-type estimate of the gradient of a smoothed loss function~\cite{stein1981estimation,nesterov2017random}. 
Likelihood-ratio and Malliavin-weight methods provide related sensitivity estimators for stochastic systems \cite{glynn1990likelihood,fournie1999applications}. 
In contrast, the proposed method reuses the noise that the NNN already needs for inference.
Instead of estimating gradients directly from scalar loss fluctuations, it estimates weights by regressing next-layer pre-activations on previous-layer fluctuations.
Errors are then propagated recursively using these weight estimates, without a separate perturbation phase for weight reconstruction.
The symmetric-probe readout estimator in Section~\ref{subsec:cov_jac_full} is also a Stein-type estimator when the probe noise is Gaussian.

Another group optimizes local objective functions using only forward propagation. 
This includes the Forward-Forward algorithm, contrasting the goodness of positive and negative examples \cite{hinton2022forward}, PEPITA utilizing an error-modulated second forward pass \cite{dellaferrera2022error}, and greedy layer-wise learning assigning auxiliary objectives to each layer \cite{nokland2019training,belilovsky2019greedy}. For spiking networks, e-prop localizes temporal credit via eligibility traces \cite{bellec2020solution}. 
All these methods offer high biological and neuromorphic plausibility due to their locality and simplicity. 
However, they suffer from fundamental issues, such as systematic errors in weight updates, poor convergence, and instability \cite{werfel2005learning,hiratani2022stability}. 
Consequently, their performance rarely matches that of backpropagation \cite{bartunov2018assessing,moskovitz2019feedback}. The proposed learning rule shares simplicity and locality with these methods by updating weights using only local statistics and treating spontaneous fluctuations of crossing activation as perturbations. 
However, its objective is to reconstruct backpropagation using local statistics, providing a new compromise between Section~\ref{subsec:2.2} and Section~\ref{subsec:2.3}.

\section{Noise-modulated Neural Network}
\label{sec:nnn}

This section formalizes the NNN examined in this study. 
Section~\ref{subsec:crossing} introduces the crossing activation function as its basic element, presenting its expected response, noise-induced local derivative, and distribution-free gradient estimation. 
Section~\ref{subsec:network} defines the network architecture stacked with these functions and its readout via ensemble averaging while introducing a noise field to control the noise of each unit.

\subsection{Crossing Activation Function}
\label{subsec:crossing}

The basic element of the NNN is the crossing activation function, which fires only in the presence of noise. 
For an input $d \in \mathbb{R}$, the output $z \in \{0,1\}$ is defined using independent and identically distributed noise $\eta_1, \eta_2 \overset{\mathrm{i.i.d.}}{\sim} p(\xi)$ as follows:
\begin{equation}
z = \phi(d) =
\begin{cases}
1 & \text{if } (d \ge \eta_1) \,\dot\vee\, (d \ge \eta_2) \\
0 & \text{otherwise}
\end{cases}
\end{equation}
where $\dot\vee$ denotes the exclusive OR (XOR) operation.                                      
Intuitively, this activation function acts as a simple spiking neuron that outputs 1 only when the input crosses the noise. 
Without noise, corresponding to $p(\xi)=\delta(\xi)$ where $\delta$ is the Dirac delta distribution concentrated at zero, no crossing event occurs for a constant input, making the output identically $0$.

The cumulative distribution function of the noise is defined as follows:
\begin{equation}
F(d) = P(d \ge \eta) = \int_{-\infty}^{d} p(\xi)\, d\xi.
\end{equation}
The expected value of the crossing activation is then given by the following equation.
\begin{equation}
\mathbb{E}[z] = \bar\phi(d) = 2F(d)\bigl(1 - F(d)\bigr).
\end{equation}

Because $F$ is a cumulative distribution function, it increases monotonically from $0$ to $1$ as $d$ increases, so the product $F(1-F)$ rises from $0$, reaches its maximum $1/4$ where $F(d) = 0.5$, and returns to $0$. Hence $\bar\phi : \mathbb{R} \to [0, 0.5]$ is a unimodal, bell-shaped response centered at the median of the noise, functioning as a smooth basis similar to a radial basis function. 
Although the binary output $z$ loses most information about the continuous input, the collective firing statistic $\bar\phi(d)$ preserves this continuous information through stochastic resonance.
Differentiating the expected response $\bar\phi$ with respect to $d$ yields the noise-induced local derivative as follows:
\begin{equation}
\bar\phi'(d) = 2\bigl(1 - 2F(d)\bigr)\, p(d).
\end{equation}
This indicates that the noise-induced statistical gradient is analytically determined, despite the system consisting of binary step functions. 
The derivative $\bar\phi'$ is positive below the point where $F(d) < 0.5$, negative above it, and vanishes away from the point. 
As discussed in Section~\ref{subsec:2.1}, while surrogate gradients in spiking networks replace step-function derivatives with hand-designed smooth functions, the NNN derives both the forward response and its derivative from the same noise statistics, eliminating any mismatch between inference and learning \cite{ikemoto2021noise}.

The crossing activation function does not restrict the noise distribution.
The definition of $\phi$, the expected response $\bar\phi = 2F(1 - F)$, and the local derivative $\bar\phi' = 2(1 - 2F)\,p$ hold directly for any $p(\xi)$ with a cumulative distribution function $F$.
For example, in the case of Gaussian noise $p(\xi) = \mathcal{N}(0, \sigma^2)$, $F(d) = \frac{1}{2}\bigl(1 + \operatorname{erf}\frac{d}{\sqrt{2}\,\sigma}\bigr)$, the expected response and local derivative are given as follows:
\begin{eqnarray}
\bar\phi(d) &=& \frac{1}{2}\left(1 - \operatorname{erf}^{2}\!\frac{d}{\sqrt{2}\,\sigma}\right) \\
\bar\phi'(d) &=& -\sqrt{\frac{2}{\pi}}\,\frac{1}{\sigma}\,\operatorname{erf}\!\left(\frac{d}{\sqrt{2}\,\sigma}\right) e^{-d^{2}/(2\sigma^{2})}
\end{eqnarray}
where $\operatorname{erf}$ denotes the error function.
The experiments in this paper primarily use this Gaussian noise.

Although the analytical form of $\bar\phi'$ requires prior knowledge of the noise distribution $p(\xi)$, the crossing activation embeds a distribution-free derivative estimator.
Consider two crossing outputs $z^{\pm} = \phi(d \pm h)$ shifted by $\pm h$ under the same noise sample, with their respective $T$-sample averages denoted as $\bar z^{+}$ and $\bar z^{-}$. 
Using these averages, the estimators for the expected response and its local derivative are obtained as follows:
\begin{eqnarray}
\phi_T(d) &=& \frac{1}{2}\bigl(\bar z^{+} + \bar z^{-}\bigr) \approx \bar\phi(d) \\
\phi_T'(d) &=& \frac{\bar z^{+} - \bar z^{-}}{2h} \approx \bar\phi'(d)
\end{eqnarray}
The estimator $\phi_T'$ corresponds to a kernel density estimation using a uniform kernel with bandwidth $h$, converging to $\bar\phi'$ as $T \to \infty$ and $h \to 0$. 
Statistically, it is crucial that $z^{+}$ and $z^{-}$ are evaluated on the same noise sample. 
Shifting the threshold by $\pm h$ is equivalent to shifting the pre-activation $d$ by $\mp h$. 
Therefore, $\phi_T'$ represents a symmetric finite difference $[z(d+h) - z(d-h)]/(2h)$ under common random numbers, which exhibits low variance because the shared noise cancels out through subtraction. 
Consequently, the crossing activation embeds a gradient estimator that performs variance reduction of the antithetic or common random numbers type directly on its own samples without additional costs. 
This slope estimation never references $p(\xi)$, allowing it to operate with the same circuit and the same code for arbitrary noise distributions.
The forward learning rules in Section~\ref{sec:forward_only_learning_rule} employ this $\phi_T'$ as the local sensitivity $\partial z / \partial d$. 

Unless otherwise stated, learning rules that require a local activation derivative use the estimator $\phi_T'$. 
This includes the backpropagation reference, which uses $\phi_T'$ as the backward derivative of the crossing activation. 
The analytical derivative $\bar\phi'$ is used for comparison in the ablation of Section~\ref{subsec:ablation} and in the closed-form expressions of Section~\ref{subsec:noise_selection}.

The derivation above assumes a continuous noise density $p(\xi)$. 
Noise quantized to $b$ bits instead takes at most $2^b$ distinct values. For two independent draws from this discrete distribution, the expected response remains $\mathbb{E}[z]=2F(1-F)$, where $F$ is the discrete cumulative distribution function.
This response is a staircase, with zero derivative between jumps and no derivative at the jumps. 
The estimator $\phi_T'$, however, remains well defined. It estimates the central finite difference of the expected response from crossing counts at $d+h$ and $d-h$. 
Section~\ref{subsec:noise_selection} evaluates the effect of bit depth on learning performance.

\subsection{Network Structure and Readout}
\label{subsec:network}

The NNN examined in this study is a fully-connected feedforward network that applies crossing activation element-wise. 
For an input vector $\mathbf{z}^{(0)} = \mathbf{x}$, the pre-activation and activation of hidden layer $l = 1, \dots, L-1$ are defined as follows:
\begin{eqnarray}
\mathbf{d}^{(l)} &=& W^{(l)} \mathbf{z}^{(l-1)} + \mathbf{b}^{(l)} \\
\mathbf{z}^{(l)} &=& \Phi^{(l)}\!\bigl(\mathbf{d}^{(l)}\bigr)
\end{eqnarray}
where $\Phi^{(l)}$ represents the element-wise application of $\phi$. 
The output layer uses a linear readout as follows:
\begin{equation}
\mathbf{y} = W^{(L)} \mathbf{z}^{(L-1)} + \mathbf{b}^{(L)}
\end{equation}
Trainable parameters are restricted to $\theta = \{W^{(l)}, \mathbf{b}^{(l)}\}$, and noise is a hyperparameter.
Regarding notation, lowercase bold letters denote vectors listing the quantities of all units within a layer, and uppercase $W^{(l)}$ denotes the connection weight matrix.

As detailed in \cite{ikemoto2026spatial}, the crossing activation has three implementation levels. 
These are the sample level ($\phi$ itself) outputting binary stochastic spikes, the statistical level outputting stochastic continuous values via $T$-sample averages, and the analytical level deterministically outputting the expected value $\bar\phi$. 
This study focuses on the sample-level network.

In a single inference pass, each unit independently samples noise $T$ times to generate a sequence of binary firing samples $z^{(l)}_{(m)}\ (m = 1, \dots, T)$.
The network output is the ensemble average of the linear readout, as follows:
\begin{equation}
\bar{\mathbf{y}} = \frac{1}{T} \sum_{m=1}^{T} \mathbf{y}_{(m)}
\end{equation}
Crucially, these $T$ stochastic forward samples are originally computed for inference. 
The learning rules in Section~\ref{sec:forward_only_learning_rule} reuse them as statistics for credit assignment without requiring additional stochastic forward passes.
The sample-wise loss $L_{(m)} = \|\mathbf{y}_{(m)} - \mathbf{t}\|^2$ and the firing fluctuation of each unit $z_{(m)} - \bar z$ serve as raw data for the covariance estimation in Section~\ref{sec:forward_only_learning_rule}.

This formulation clearly shows that the NNN can be trained via backpropagation. 
The network is a composition of layers with a linear readout, and each crossing activation provides its expected response $\bar\phi$ and its local derivative, such as the analytical form $\bar\phi'$ or the distribution-free estimator $\phi_T'$, during forward propagation. 
Therefore, standard backpropagation applies directly by propagating the readout error $2(\bar{\mathbf{y}} - \mathbf{t})$ backward along the chain rule, multiplying by $W^{(l)\mathsf{T}}$ and the local derivative at each layer, enabling efficient training of the NNN. 
This study starts from this baseline that relies on weight transport via the backward path using the transposed forward weights $W^{(l)\mathsf{T}}$. 
Section 4 reconstructs these transposed weights and the backward path solely from statistics obtained during forward propagation.

It is noteworthy that the noise distribution does not need to be shared across all units. 
Letting $p^{(l,k)}$ be the noise distribution of the $k$-th unit in layer $l$, the entire set is termed the noise field. 
The noise field is given by the following expression:
\begin{equation}
\mathcal{P} = \bigl\{\, p^{(l,k)} \,\bigr\}_{l=1 \dots L-1,\ k = 1 \dots K(l)}
\end{equation}
where $K(l)$ indicates the number of units in the $l$-th layer.
Units with $p^{(l,k)} = \delta$ (no noise) produce outputs and gradients that are identically $0$, completely disconnecting them from both inference and learning. 
Thus, the noise field acts as a hyperparameter that determines which sub-network is recruited for computation.
This study employs a minimal configuration where noise intensity (scale parameter, such as $\sigma$ in a Gaussian distribution) in each unit is modulated by a scalar field $s_{k} \in [0, 1]$.
The learning rules in Section~\ref{sec:forward_only_learning_rule} naturally align with this field, meaning units with $s_k = 0$ receive a covariance credit of exactly $0$ and are automatically excluded from updates.
This implies that credit assignment and recruitment share the same field, which has implications for the power gating discussed in Section~\ref{subsec:noise_selection}.

\section{Forward Fluctuation Covariance Learning}
\label{sec:forward_only_learning_rule}

The NNN in Section~\ref{sec:nnn} generates multiple stochastic forward samples for each input, including layer-wise pre-activations, activations, readout outputs, and sample-wise losses. 
The learning rules proposed in this section estimate the gradients for weight updates using only these forward quantities.
They require no transpose access to the forward weights or autograd. 
Because optimizers such as SGD and Adam \cite{kingma2015adam} operate locally without causing weight transport, estimating gradients from forward statistics is the core requirement to solve the weight transport problem. 
Following the problem setup in Section~\ref{subsec:problem_setup}, I sequentially introduce the simplest scalar covariance rule in Section~\ref{subsec:cov_deriv}, the structured covariance rule in Section~\ref{subsec:cov_jac} to overcome its limitations, and the generalization in Section~\ref{subsec:cov_jac_full} to replace the remaining analytical loss derivatives.

\subsection{Problem Setup}
\label{subsec:problem_setup}

For a dataset $\mathcal{D} = \{\mathbf{x}_n, \mathbf{t}_n\}_{n=1}^{N}$, the NNN in Section~\ref{subsec:network} generates $T$ stochastic forward samples per input. 
The network consists of layers $l = 1, \dots, L$, where layer $L$ is a linear readout.
Let $\mathbf{d}^{(l)}_{(m,n)}$ and $\mathbf{z}^{(l)}_{(m,n)}$ be the pre-activation and activation of layer $l$ for input $n$ and sample $m$, $\mathbf{y}_{(m,n)}$ be the readout output, and $L_{(m,n)} = \|\mathbf{y}_{(m,n)} - \mathbf{t}_n\|^2$ be the sample-wise loss. 
Subscripts in parentheses $(m,n)$ denote the sample and input indices, whereas subscripts without parentheses $i$ and $j$ indicate vector elements or units. 
For example, $z^{(l)}_{(m,n),i}$ is the $i$-th element of $\mathbf{z}^{(l)}_{(m,n)}$, representing the activation of unit $i$ in layer $l$. 
Regarding notation, an overbar denotes a $T$-sample average for a fixed input $n$, and omitting the sample index $m$ refers to an observation sequence over the same $T$ samples. 
The operator $\langle \cdot \rangle_{m,n}$ represents the average over all samples and inputs, and $\epsilon$ is a small positive constant to prevent division by zero. 
The network output is the ensemble average $\bar{\mathbf{y}}_n$. 
The training objective function $L$ is the average squared error $\|\bar{\mathbf{y}}_n - \mathbf{t}_n\|^2$ over all inputs, which is contextually distinguished from the layer index $L$. The readout error signal is given by $\mathbf{e}_n = 2(\bar{\mathbf{y}}_n - \mathbf{t}_n)$.

Because $\bar{\mathbf{y}}_n$ is linear with respect to $W^{(L)}$, the readout gradient can be expressed exactly and locally using the ensemble-averaged activation $\bar{\mathbf{z}}^{(L-1)}_n$ as follows:
\begin{equation}
\frac{\partial L}{\partial W^{(L)}} = \frac{1}{N} \sum_{n} \mathbf{e}_n\, \bar{\mathbf{z}}^{(L-1)\mathsf{T}}_n
\end{equation}
Therefore, eliminating the direct access to the transposed forward weights reduces entirely to the hidden-layer credit assignment problem, which requires estimating $\partial L / \partial z^{(l)}_i$ solely from forward statistics.

\subsection{Scalar Covariance Credit Rule}
\label{subsec:cov_deriv}

The simplest configuration is the regression of the loss on the spontaneous fluctuations of each unit. 
Using the covariance centered over $T$ samples for each input $n$, the credit for the activation is defined as follows:
\begin{equation}
g^{(l)}_{n,i} = \frac{\mathrm{Cov}_T\bigl(L_n,\, z^{(l)}_{n,i}\bigr)}{\mathrm{Var}_T\bigl(z^{(l)}_{n,i}\bigr) + \epsilon}
\;\approx\; \frac{\partial L_n}{\partial z^{(l)}_{n,i}}.
\end{equation}
Here, $\mathrm{Cov}_T$ and $\mathrm{Var}_T$ are the empirical covariance and variance over $T$ samples for a fixed input $n$.
These terms are calculated from only four sample averages accumulated during the forward pass, specifically $\overline{L}_n$, $\overline{z}_{n,i}$, $\overline{Lz}_{n,i}$, and $\overline{z^2}_{n,i}$, which represent the averages of $L$, $z_i$, $L z_i$, and $z_i^2$, respectively, with the layer index omitted as follows:
\begin{eqnarray}
\mathrm{Cov}_T\bigl(L_n,\, z^{(l)}_{n,i}\bigr) &=& \overline{Lz}_{n,i} - \overline{L}_n\, \overline{z}_{n,i} \\
\mathrm{Var}_T\bigl(z^{(l)}_{n,i}\bigr) &=& \overline{z^2}_{n,i} - \overline{z}_{n,i}^{\,2}.
\end{eqnarray}
This credit replaces the $W^{(l+1)\mathsf{T}} \boldsymbol{\delta}^{(l+1)}$ term of backpropagation, where $\boldsymbol{\delta}^{(l+1)}$ is the error signal propagated from higher layers, with forward statistics.

Multiplying this credit by the distribution-free local slope $\phi_T'$ from Section 3.1 constructs a pseudo-error, yielding the hidden-layer weight gradient estimate as follows:
\begin{eqnarray}
\hat\delta^{(l)}_{n,i} &=& g^{(l)}_{n,i}\, \phi_T'\bigl(d^{(l)}_{n,i}\bigr) \\
\hat{G}^{(l)} &=& \Bigl\langle \hat{\boldsymbol\delta}^{(l)} \mathbf{z}^{(l-1)\mathsf{T}} \Bigr\rangle_{m,n}
\;\approx\; \frac{\partial L}{\partial W^{(l)}}
\end{eqnarray}
Because $g \cdot \phi_T' \approx (\partial L/\partial z)(\partial z/\partial d) = \partial L/\partial d$, the gradient estimate takes the same outer-product form $\boldsymbol{\delta} \mathbf{z}^{\mathsf{T}}$ as backpropagation. 
This rule, which forms the pseudo-error $\hat{\boldsymbol\delta}$ by multiplying the scalar covariance credit $g$ by the local slope $\phi_T'$ and computes $\hat{G}^{(l)}$ via an outer product, is referred to as ``\texttt{cov\_deriv}'' in this paper.

The \texttt{cov\_deriv} rule can be interpreted as deeply integrating node perturbation into the NNN. 
Classical node perturbation injects an independent perturbation $\xi$ into each unit solely for training and regresses the loss change on $\xi$. 
In contrast, because crossing activation already fluctuates due to the forward noise, the spontaneous fluctuation $z - \bar{z}$ itself acts as the perturbation, eliminating the need for an additional dedicated process.
As a related simple configuration, a minimal variant ``\texttt{cov\_only}'' can be considered, which omits the local slope $\phi_T'$ and directly uses the credit for the gradient estimate. These covariance statistics require only the four accumulations per unit ($\overline{L}$, $\overline{z}$, $\overline{Lz}$, and $\overline{z^2}$), while \texttt{cov\_deriv} additionally employs a crossing counter for $\phi_T'$. Both configurations allow online accumulation as samples are obtained.

However, both \texttt{cov\_only} and \texttt{cov\_deriv} face fundamental concerns compared to backpropagation. 
First, \texttt{cov\_only} lacks the $\partial z / \partial d$ information as a trade-off for its simplification, which is expected to distort the update direction. 
Second, the credit estimator of \texttt{cov\_deriv} retains a systematic error inherent in its structure. 
The ratio $\mathrm{Cov}(L, z_i)/\mathrm{Var}(z_i)$ is a univariate regression that compresses the downstream network effects of unit $i$ into the scalar loss $L$, linearizing and ignoring interactions when multiple units fluctuate simultaneously. 
Because this linearization error is not a finite-sample effect, it does not vanish by increasing $T$. 
Avoiding this compression requires computing the covariance with quantities aligned with the computational graph structure rather than the scalar $L$.

\subsection{Structured Covariance Credit Rule}
\label{subsec:cov_jac}

\subsubsection{Bypassing the Weight Transport Problem}
\label{subsubsec:bypassing}

Pre-activations are strictly linear with respect to activations ($\mathbf{d}^{(l+1)} = W^{(l+1)} \mathbf{z}^{(l)} + \mathbf{b}^{(l+1)}$). 
Therefore, the pooled single-regression coefficients of the $T$-sample fluctuations for all inputs, when the input is fixed, yield the forward weights themselves as follows:
\begin{equation}
\widehat W^{(l+1)}_{ji}
= \frac{\sum_{n} \mathrm{Cov}_T\bigl(d^{(l+1)}_{n,j},\, z^{(l)}_{n,i}\bigr)}{\sum_{n} \mathrm{Var}_T\bigl(z^{(l)}_{n,i}\bigr) + \epsilon}
\;\approx\; W^{(l+1)}_{ji}
\end{equation}

This approach adapts the weight-mirror method, which estimates the transposed weight matrix from input-output correlations under noise injection \cite{akrout2019deep}. 
The key difference is that instead of requiring a dedicated noise injection phase, this approach exploits the fluctuations already generated by the NNN during the forward pass. 
In addition to the accumulated quantities in Section~\ref{subsec:cov_deriv}, the system stores $\overline{d}_{n,j}$ for each unit and the average product $\overline{dz}_{n,ji}$ for each unit pair $(j,i)$. 
The numerator is given by $\mathrm{Cov}_T\bigl(d^{(l+1)}_{n,j},\, z^{(l)}_{n,i}\bigr) = \overline{dz}_{n,ji} - \overline{d}_{n,j}\, \overline{z}_{n,i}$, and the denominator is computed from $\overline{z}$ and $\overline{z^2}$ shared with Section~\ref{subsec:cov_deriv}. 

The following three points are key implementation requirements to improve estimation efficiency and accuracy:
\begin{enumerate}
    \item The correlation is computed with the pre-activation $d^{(l+1)}$. Since $d^{(l+1)}_j$ is the weighted sum of the previous layer firing and is strictly linear with respect to $\mathbf{z}^{(l)}$, the regression coefficient directly provides $W_{ji}$.
    \item Unlike gradients, weights do not depend on the input, so first sum the numerator and denominator over all inputs and then divide. This allows all $NT$ samples included in a single parameter update step to be used in a single estimation (resulting in lower variance than calculating the ratio for each input and then averaging it).
    \item The estimation $\widehat W$ is performed at each update step using an exponential moving average with a smoothing constant of 0.9. Weight changes are incorporated separately via $\widehat W \leftarrow \widehat W + \Delta W$, adapting the Kolen-Pollack method~\cite{kolen1994backpropagation}.
\end{enumerate}

\subsubsection{Recursive Error Propagation}
\label{subsubsec:recursive}

Using $\widehat W$ and $\phi_T'$ estimated from forward perturbation samples, the output error can be propagated through a recursion isomorphic to backpropagation.
\begin{eqnarray}
\hat{\boldsymbol\delta}^{(L-1)}_n
&=& \Bigl(\widehat W^{(L)\mathsf{T}}\, \mathbf{e}_n\Bigr) \odot \phi_T'\bigl(\mathbf{d}^{(L-1)}_n\bigr) \\
\hat{\boldsymbol\delta}^{(l)}_n
&=& \Bigl(\widehat W^{(l+1)\mathsf{T}} \hat{\boldsymbol\delta}^{(l+1)}_n\Bigr) \odot \phi_T'\bigl(\mathbf{d}^{(l)}_n\bigr)
\end{eqnarray}
The gradient estimate is obtained via the outer product $\hat{G}^{(l)} = \langle \hat{\boldsymbol\delta}^{(l)} \mathbf{z}^{(l-1)\mathsf{T}} \rangle_{m,n}$ as in \texttt{cov\_deriv}. 
Compared to backpropagation, the only difference is that $W^{\mathsf{T}}$ is replaced by $\widehat W^{\mathsf{T}}$ estimated from forward perturbations.

The inter-layer Jacobian multiplied in backpropagation via the chain rule consists of two factors, namely the linear combination part $\partial \mathbf{d}^{(l+1)} / \partial \mathbf{z}^{(l)} = W^{(l+1)}$ and the local derivative of the activation $\partial z / \partial d$. 
The recursion above replaces these two factors with $\widehat W$ and $\phi_T'$ estimated from forward statistics. 
This represents a reconstruction of backpropagation based on Jacobian estimation via forward perturbations. 
Hereafter, this method is referred to as ``\texttt{cov\_jac}.''

Unlike \texttt{cov\_deriv}, \texttt{cov\_jac} propagates the credit along the computation graph without compressing it into the scalar loss $L$. 
For a layer width of $K$ units, the statistics require the pairwise average product $\overline{dz}$, which increases the computational complexity from $O(K)$ in \texttt{cov\_deriv} to $O(K^2)$ in \texttt{cov\_jac}. 
Nevertheless, it remains forward-only and supports online accumulation. 
The simple regression for $\widehat W$ is a diagonal approximation assuming uncorrelated activations within a layer. 
Strong intra-layer correlations could cause weight leakage from neighboring units, creating a directional bias that persists even with larger sample sizes. 
However, this effect is negligible in the present regime because the unique crossing noise of each unit dominates, as quantified in Section~\ref{subsec:correlation}. This correlation bias grows with depth, as shown in Section~\ref{subsec:deep}.

\subsection{Generalized Structured Covariance Credit Rule}
\label{subsec:cov_jac_full}

Up to this point, the only analytically computed quantity is the readout error signal $\mathbf{e}_n = 2(\bar{\mathbf{y}}_n - \mathbf{t}_n)$. 
While this term is independent of weight transport, this final loss derivative can also be replaced with forward statistics. 
In the following, each readout unit is treated element-wise, omitting element subscripts to write $y$, $t$, and $e$. 
Because the sample output $y_{(m,n)}$ is a continuous quantity that fluctuates with forward perturbations, the same per-input regression used for hidden layers can be applied as follows:
\begin{equation}
g_{y,n} = \frac{\mathrm{Cov}_T\bigl(L_n,\, y_n\bigr)}{\mathrm{Var}_T\bigl(y_n\bigr) + \epsilon}
\;\approx\; e_n.
\label{eq:readout_reg}
\end{equation}
The only newly accumulated quantities are $\overline{y}_n$, $\overline{Ly}_n$, and $\overline{y^2}_n$ for each readout unit, while $\overline{L}_n$ is shared with Section~\ref{subsec:cov_deriv}. 
In this paper, the rule that uses this $g_{y,n}$ both as the starting point for recursion to seed $\hat{\boldsymbol\delta}^{(L-1)}$ and for estimating the readout weight gradient is called \texttt{cov\_jac\_full}. 
In this configuration, the network observes only the scalar loss per sample, meaning no analytical loss derivative appears anywhere.

However, this regression introduces a systematic error. 
Since the loss $L = (y - t)^2$ is a quadratic function of $y$, the population regression coefficient is strictly given by the following expression:
\begin{eqnarray}
\frac{\mathrm{Cov}(L, y)}{\mathrm{Var}(y)} &=& 2\bigl(\mathbb{E}[y] - t\bigr) + \frac{\mathbb{E}[\varepsilon^3]}{\mathrm{Var}(\varepsilon)} \\
\varepsilon &=& y - \mathbb{E}[y].
\end{eqnarray}
The second term is a bias proportional to the skewness of the readout fluctuations. 
Because $y$ is a weighted sum of binary firings, this term is generally non-zero and persists even with larger sample sizes $T$.
This bias is less prominent in SGD because the step size shrinks along with the gradient magnitude. 
However, Adam continually amplifies this bias through gradient scale normalization, which can cause the solution to drift in the later stages of training.

Two correction methods can eliminate the need for analytical loss derivatives. 
The third central moment correction subtracts the third central moment of $y$, given by $m_3 = \overline{y^3}_n - 3\,\overline{y}_n\, \overline{y^2}_n + 2\,\overline{y}_n^{\,3}$, from the numerator, where $g_y = (\mathrm{Cov} - m_3)/\mathrm{Var}$. 
The symmetric probe adds a known symmetric noise $\xi$ satisfying $\mathbb{E}[\xi^3] = 0$ to the readout samples and regresses the loss on $\xi$ instead of $y$, where $g_y = \mathrm{Cov}_T(L(y+\xi), \xi)/\mathrm{Var}_T(\xi)$. 
These two methods form a complementary pair. 
The third central moment correction requires one additional accumulation $\overline{y^3}_n$ and achieves exact bias removal only for quadratic losses. 
Nevertheless, because it operates solely on the observed quantities $(L, y)$, it requires no additional perturbation injection and integrates naturally into the NNN. 
In contrast, the symmetric probe accumulates $\overline{\xi}$, $\overline{L\xi}$, and $\overline{\xi^2}$ instead of $\overline{y}$, $\overline{Ly}$, and $\overline{y^2}$, so no additional accumulation is needed. 
However, it requires injecting an additional perturbation, which partially undermines my claim that the method integrates naturally into the NNN. 
This study uses the third central moment correction as the default configuration for the single-output quadratic loss of Section~\ref{sec:experiment}.

With $C$ readout units, the sample outputs fluctuate jointly because they share the same hidden samples. The univariate regression of $L$ on each $y_k$ then suffers from the same omitted-variable bias as the univariate weight mirror of Section~\ref{subsec:cov_jac}.
The multi-output generalization is the multivariate regression
\begin{equation}
\mathbf{g}_{y,n} = \mathrm{Cov}_T(\mathbf{y}_n, \mathbf{y}_n)^{-1}\,\mathrm{Cov}_T(\mathbf{y}_n, L_n),
\label{eq:mv_readout}
\end{equation}
which requires one $C \times C$ linear solve per input. For the squared loss $L = \|\mathbf{y} - \mathbf{t}\|^2$, the estimator becomes exact in the population limit after subtracting the vector third moment $\mathbb{E}[\|\boldsymbol\varepsilon\|^2 \boldsymbol\varepsilon]$ from $\mathrm{Cov}_T(\mathbf{y}_n, L_n)$, provided that the covariance matrix is invertible and no regularization is added.
The third-moment correction is exact only for quadratic losses, whereas the symmetric probe with Gaussian probe noise is a Stein-type estimator. 
When $\xi \sim \mathcal{N}(0, \alpha^2)$ is independent of $y$, the quantity $\mathrm{Cov}(L(y+\xi), \xi)/\alpha^2$ equals the expectation of $L'(y+\xi)$ over both $\xi$ and the fluctuation of $y$ for any smooth $L$ \cite{stein1981estimation}. This is the gradient of the smoothed loss rather than the gradient at the mean readout.
For a non-quadratic loss such as cross-entropy, the probe therefore applies where the third-moment correction does not.
Section~\ref{subsec:deep} shows that the correction fails in that setting while the probe and the multivariate regression achieve test accuracy comparable to that obtained with the analytical readout gradient.

\section{Experimental Results}
\label{sec:experiment}

\subsection{Experimental Setup}
\label{subsec:setup}

This section evaluates a simple regression task approximating $\sin(x)$. 
The network architecture is 1–64–64–1 with an ensemble size of $T=64$, and training is performed for 1500 epochs using identical initial weights across all configurations. 
Unless stated otherwise, the Gaussian noise has $\sigma = 0.5$, the slope estimator uses $h = 0.2$, the regularizer is $\epsilon = 10^{-6}$, and the learning rate is $10^{-2}$.
The evaluation compares five methods, namely backprop, \texttt{cov\_only}, \texttt{cov\_deriv}, \texttt{cov\_jac}, and \texttt{cov\_jac\_full}. 
The four covariance-based learning rules introduced in Section~\ref{sec:forward_only_learning_rule} operate without autograd, updating weights manually under the no\_grad context. 
The credit and computational requirements for each method are summarized in Table~\ref{tab:methods}.

\begin{table}[tb]
\centering
\caption{Comparison of the learning signals and computational requirements of the evaluated methods. The table lists the hidden-layer credit, readout-error signal, optimizer, asymptotic additional memory per layer, additional computation per parameter update, and the requirement for a backward path or transposed-weight access. Here, $H$ denotes the hidden-layer width and $T$ denotes the number of stochastic forward samples. The corresponding learning rules are detailed in Section~\ref{sec:forward_only_learning_rule}.}
\label{tab:methods}
\renewcommand{\tabularxcolumn}[1]{m{#1}}
\begin{tabularx}{0.96\textwidth}{*{6}{>{\raggedright\arraybackslash}X}}
\hline
Method & backprop & \texttt{cov\_only} & \texttt{cov\_deriv} & \texttt{cov\_jac} & \texttt{cov\_jac\_full} \\
\hline
Hidden Credit & $W^{\mathsf{T}}\boldsymbol{\delta}$ & $\frac{\mathrm{Cov}(L,z)}{\mathrm{Var}(z)}$ & $\frac{\mathrm{Cov}(L,z)}{\mathrm{Var}(z)} \phi_T'$ & $\widehat{W}^{\mathsf{T}}\hat{\boldsymbol{\delta}}$ & $\widehat{W}^{\mathsf{T}}\hat{\boldsymbol{\delta}}$ \\
\hline
Readout Error & $2(\bar{y}-t)$ & $2(\bar{y}-t)$ & $2(\bar{y}-t)$ & $2(\bar{y}-t)$ & $\frac{\mathrm{Cov}(L,y)-m_3}{\mathrm{Var}(y)}$\\
\hline
Optimizer & Adam & SGD & SGD & Adam & Adam \\
\hline
Extra Memory per Layer & $O(H^2)$ & $O(H)$ & $O(H)$ & $O(H^2)$ & $O(H^2)$ \\
\hline
Extra Compute per Update & $O(H^2)$ & $O(TH)$ & $O(TH)$ & $O(TH^2)$ & $O(TH^2)$ \\
\hline
Access to Transposed Weights & Required & Not required & Not required & Not required & Not required \\
\hline
\end{tabularx}
\end{table}

Section~\ref{sec:forward_only_learning_rule} describes the hidden-layer credit and readout error, allowing parameter updates with any optimizer. 
In addition to SGD, the first and second moments $m$ and $v$ of Adam are local to each weight, ensuring that Adam causes no weight transport. 
However, the suitability of the optimizer depends on the error properties of the estimator. 
The $1/\sqrt{v}$ normalization of Adam standardizes the gradient scale. 
When applied to \texttt{cov\_only} or \texttt{cov\_deriv}, which have the persistent systematic errors described in Section~\ref{subsec:cov_deriv}, this normalization continuously amplifies the bias component as the true gradient shrinks in later training stages, pushing the solution away from the optimum. 
In contrast, Adam operates effectively for the low-variance, quasi-unbiased gradients of \texttt{cov\_jac} and \texttt{cov\_jac\_full} as in backpropagation. 
Therefore, the main evaluation adopts the specific optimizer suited for each learning rule, as shown in Table~\ref{tab:methods}.

\subsection{Main Results}
\label{subsec:main_results}

\begin{figure}[tb]
 \centering
    \includegraphics[width=1.0\textwidth]{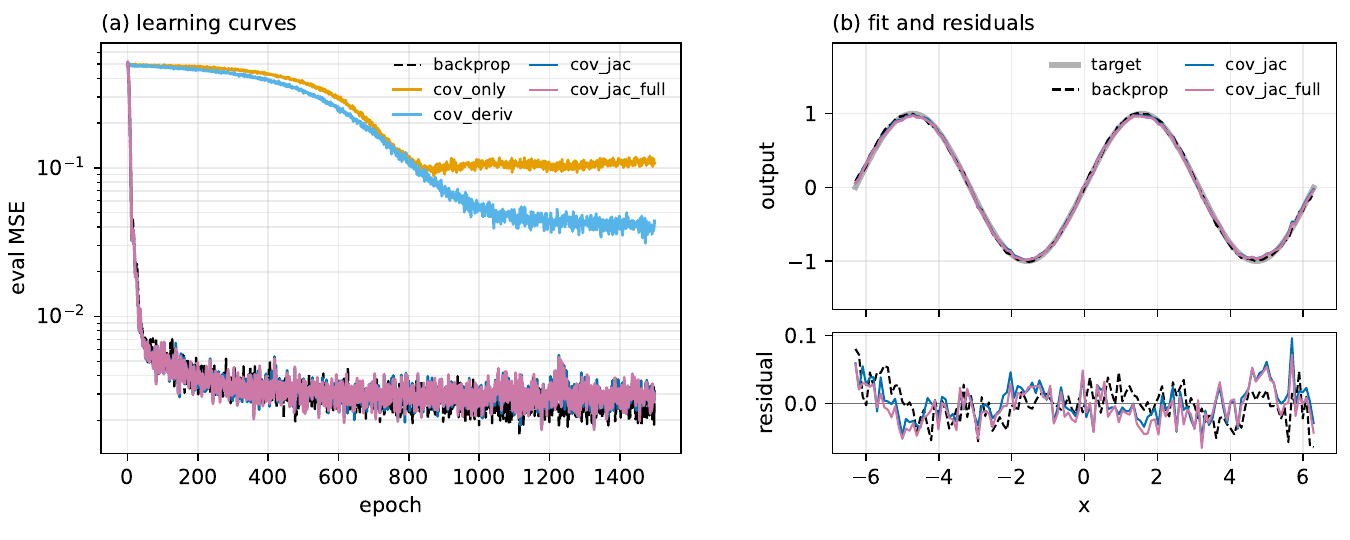}
 \caption{Learning performance on the sin(x) regression task using backprop and proposed method variants. (a) Evaluation MSE over 1500 training epochs for backprop, \texttt{cov\_only}, \texttt{cov\_deriv}, \texttt{cov\_jac}, and \texttt{cov\_jac\_full}. (b) Target function, learned outputs, and residuals after training for backprop, \texttt{cov\_jac}, and \texttt{cov\_jac\_full}.}
\label{fig:main_result}
\end{figure}

\begin{table}[tb]
\centering
\caption{Final evaluation MSE on the $\sin(x)$ regression task after 1500 training epochs. Values are reported as the mean $\pm$ standard deviation over three different seeds for the 1–64–64–1 network with $T=64$ stochastic forward samples per input.}
\label{tab:performance}
\begin{tabular}{ll}
\hline
Method & Final MSE \\
\hline
backprop & $0.00057 \pm 0.00010$ \\
\texttt{cov\_only} & $0.09583 \pm 0.00724$ \\
\texttt{cov\_deriv} & $0.03464 \pm 0.00558$ \\
\texttt{cov\_jac} & $0.00056 \pm 0.00006$ \\
\texttt{cov\_jac\_full} & $0.00057 \pm 0.00009$ \\
\hline
\end{tabular}
\end{table}

Table~\ref{tab:performance} lists the final MSE (mean $\pm$ std over seeds 0--2) for each method, and Fig.~\ref{fig:main_result} shows the training curves and post-training responses for the first seed.
The final MSE values of \texttt{cov\_jac} and \texttt{cov\_jac\_full} matched backprop within the seed-to-seed variability. In contrast, \texttt{cov\_only} and \texttt{cov\_deriv} using scalar covariance credit were limited to $0.096$ and $0.035$, respectively. 
This residual error aligns with the persistent systematic errors described in Section~\ref{subsec:cov_deriv}.
As shown in Fig.~\ref{fig:main_result}(a), the training curves of \texttt{cov\_jac} and \texttt{cov\_jac\_full} closely overlap with backprop throughout all epochs, showing no difference in convergence speed. 
In Fig.~\ref{fig:main_result}(b), the predictions of both methods overlap with the target function, and the residuals have amplitudes comparable to backprop without exhibiting systematic structures.
These results demonstrate that structuring credits via mirror recursion (Section~\ref{subsec:cov_jac}) achieves a learning performance comparable to backpropagation without transporting or transposing weights.
Furthermore, because \texttt{cov\_jac\_full} avoids analytical loss derivatives, all signals required for training are derived from forward sampling statistics.

\subsection{Verification of the Accuracy of Gradient Reconstruction}
\label{subsec:gradient_fidelity}

\begin{figure}[tb]
 \centering
    \includegraphics[width=1.0\textwidth]{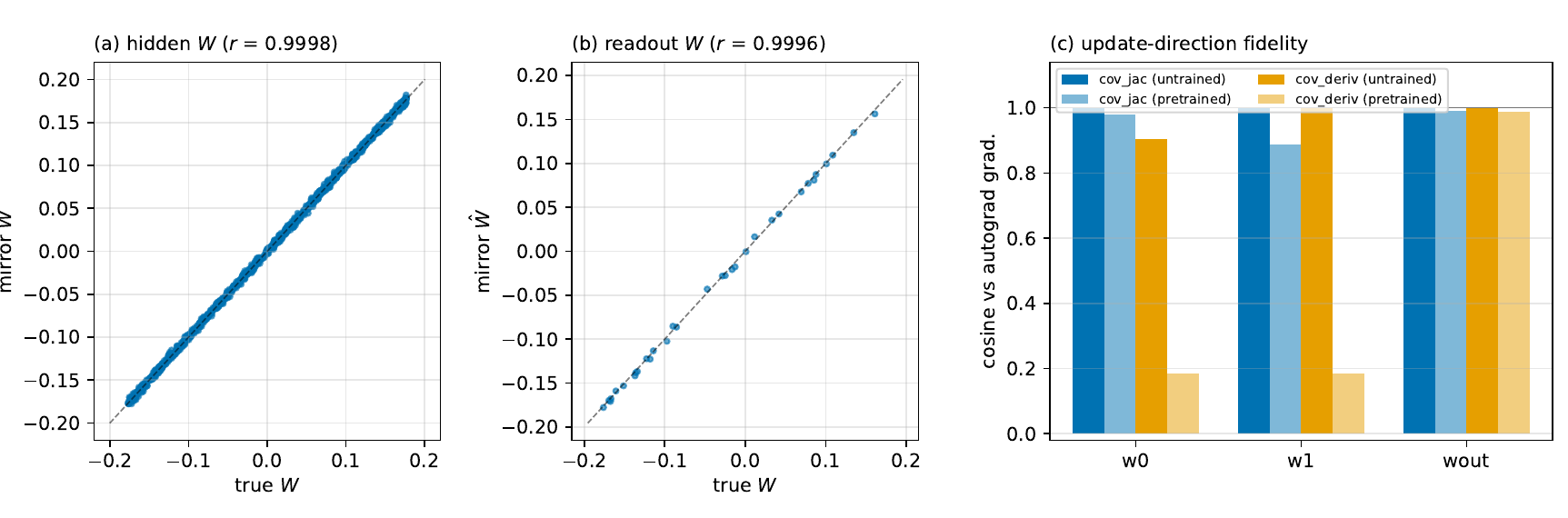}
 \caption{Fidelity of the reconstructed weights and gradient directions. (a, b) Recovered mirror weights $\widehat{W}$
 versus the true forward weights $W$ for the hidden and readout layers, respectively, in the untrained network. Here $r$ denotes the Pearson correlation coefficient. (c) Cosine similarity between the layer-wise update directions produced by \texttt{cov\_jac} or \texttt{cov\_deriv} and the exact autograd gradients at initialization and after 300 epochs of backpropagation pretraining. Here, ``w0'' and ``w1'' denote the two hidden-layer weight matrices, and ``wout'' denotes the readout weight matrix.}
\label{fig:gradient_fidelity}
\end{figure}

To confirm that the agreement in Section~\ref{subsec:main_results} is not accidental, I directly compare the learning signals of \texttt{cov\_jac} with exact reference quantities (Fig.~\ref{fig:gradient_fidelity}).
The network in this section has 32 units per hidden layer.
First, I evaluate the recovery accuracy of the weight mirror (See Section~\ref{subsec:cov_jac}), which forms the core of the recursion. 
Figs.~\ref{fig:gradient_fidelity}(a) and (b) show the scatter plots of the recovered $\widehat{W}$ and the true forward weights $W$ in the untrained state, where the values align on the same straight line for both the hidden and readout layers. 
The Pearson correlation reaches $r \geq 0.999$ in the hidden layers and $r \geq 0.988$ in the readout layer in both untrained and partially trained states after 300 epochs of backpropagation. 
Next, I compare the layer-wise updates constructed by \texttt{cov\_jac} with the exact gradients from backpropagation via autograd (Fig.~\ref{fig:gradient_fidelity}(c)). 
In the untrained initial state, the cosine similarity is 0.998 to 1.000 across all layers, and the norm ratio is 0.99 to 1.00, demonstrating that \texttt{cov\_jac} reconstructs the gradient almost exactly in both direction and scale. 
Even during training, the directional agreement is maintained with a cosine similarity of 0.89 to 0.99, although the norm ratio expands up to 2.2. 
This scale discrepancy is absorbed by the $1/\sqrt{v}$ normalization of Adam, meaning it does not appear in the final performance (Section~\ref{subsec:main_results}). 
In contrast, the update direction of \texttt{cov\_deriv} drops to a cosine similarity of 0.19 in the hidden layers during training (Fig.~\ref{fig:gradient_fidelity}(c)). 
The performance gap between \texttt{cov\_jac} and \texttt{cov\_deriv} observed in Section~\ref{subsec:main_results} is directly explained by this difference in gradient fidelity.

\subsection{Ablation Study}
\label{subsec:ablation}

\begin{figure}[tb]
 \centering
    \includegraphics[width=1.0\textwidth]{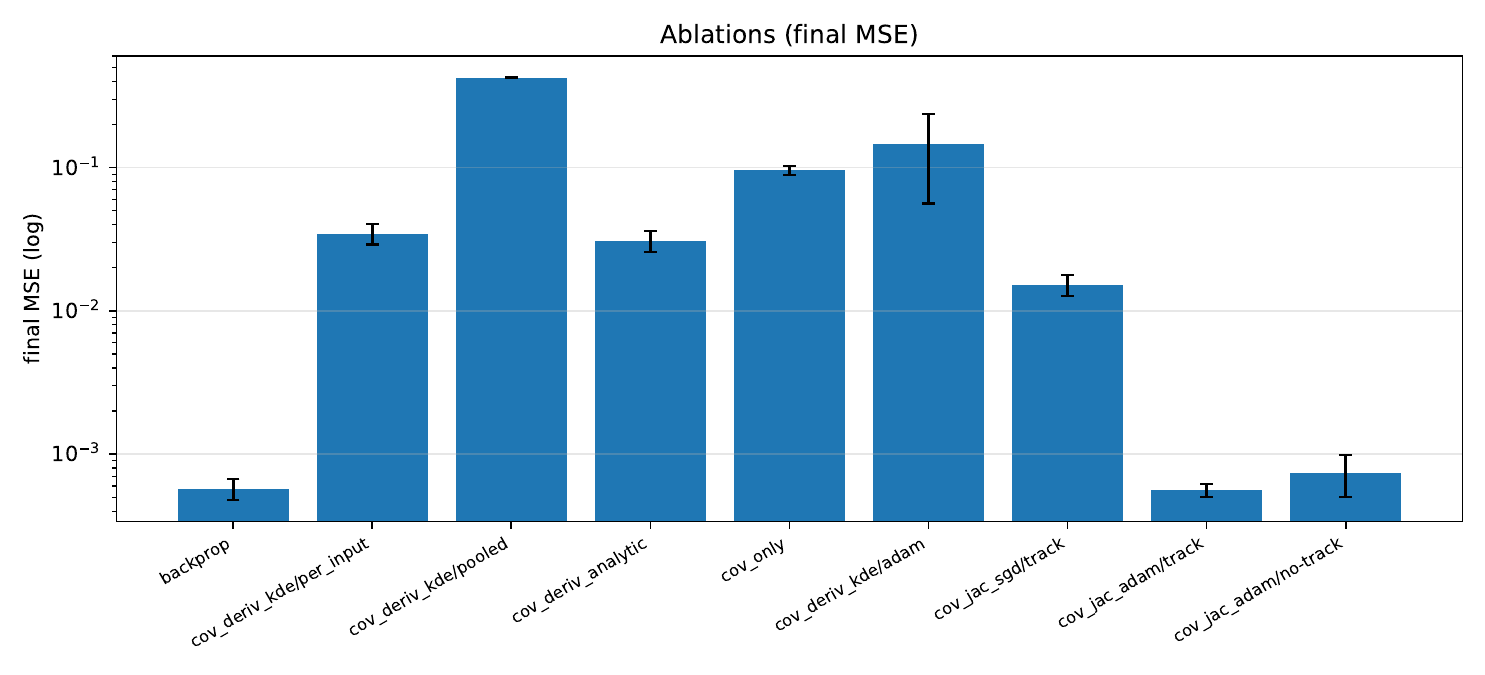}
 \caption{Ablation study of the proposed learning rules. Bars show the final evaluation MSE on a logarithmic scale, with the mean and standard deviation evaluated over three different seeds. The comparisons examine the effects of optimizer selection, per-input versus pooled covariance estimation, distribution-free versus analytical local-slope estimation, omission of the local slope, and explicit tracking of parameter updates in the covariance weight mirror. Backpropagation is included as a reference.}
\label{fig:ablation}
\end{figure}

The main results in Section~\ref{subsec:main_results} rely on a combination of design choices, including credit structuring, per-input credit estimation, distribution-free slope estimation, and an estimator-suited optimizer. 
This section isolates which choices are essential for performance and which are interchangeable.
Specifically, I alter four main design axes one by one while maintaining the protocol of Section~\ref{subsec:setup}.
These are (1) optimizer (SGD or Adam), (2) credit estimation unit (per-input or pooled), (3) local slope estimation method (estimated slope $\phi'_T$ or analytical derivative $\bar\phi'$), and (4) presence of the local slope (\texttt{cov\_deriv} or \texttt{cov\_only}). 
In addition, I examine the effect of explicitly tracking parameter updates in the covariance weight mirror. The results are summarized in Fig.~\ref{fig:ablation}.

\subsubsection{Optimizer}

When \texttt{cov\_jac} is trained with SGD, the final MSE is limited to 0.015, whereas it reaches 0.00056 with Adam, as shown by the \texttt{cov\_jac\_sgd/track} and \texttt{cov\_jac\_adam/track} conditions in Fig.~\ref{fig:ablation}, respectively.
This residual error is an artifact of optimization rather than a bias in the credit estimator. 
Conversely, switching \texttt{cov\_deriv} to Adam degrades the final MSE from 0.035 to $0.147 \pm 0.091$, and the seed-to-seed variability becomes the largest among all comparisons. 
This indicates that the systematic error described in Section~\ref{subsec:cov_deriv} is not eliminated by the optimizer, and Adam, which normalizes the gradient scale, amplifies it instead. 
This asymmetry justifies the choice of optimizers in Section~\ref{subsec:setup}, and the amplification mechanism is directly demonstrated for the readout estimator in Section~\ref{subsec:readout_error}.
Removing the explicit mirror-tracking update caused only a small degradation under the present conditions, as shown by the \texttt{cov\_jac\_adam/no-track} condition in Fig.~\ref{fig:ablation}.
In this condition, the weight mirror is estimated per input instead of being pooled over all inputs.

\subsubsection{Credit estimation unit}

Replacing the per-input credit, which computes covariance statistics for each input, with pooled credit across all inputs degrades the final MSE from 0.035 to 0.42. 
This represents the largest degradation in Fig.~\ref{fig:ablation}, and learning practically fails. 
Because the covariance between loss and activity approximates the local gradient only when conditioned on the input (Section~\ref{subsec:cov_deriv}), per-input estimation is critical.

\subsubsection{Local slope estimation method}

Replacing the distribution-free estimated slope (Section~\ref{subsec:crossing}) with the analytical $\bar\phi'$ keeps the final MSE nearly unchanged at 0.031 compared to 0.035 (Fig.~\ref{fig:ablation}). 
Therefore, the two are interchangeable, and an analytical model of the noise distribution is unnecessary for learning.

\subsubsection{Presence of local slope}

The difference between \texttt{cov\_only} without $\bar\phi'$ (0.096) and \texttt{cov\_deriv} with it (0.035) illustrates the contribution of multiplying the covariance credit by the local slope (Fig.~\ref{fig:ablation}).
In conclusion, performance is determined by credit structuring (Section~\ref{subsec:main_results}), per-input estimation, and an optimizer suited to the error properties, while the implementation of slope estimation is interchangeable.

\subsection{Empirical Study on the Estimation of the Covariance of Readout Errors}
\label{subsec:readout_error}

\begin{figure}[tb]
 \centering
    \includegraphics[width=1.0\textwidth]{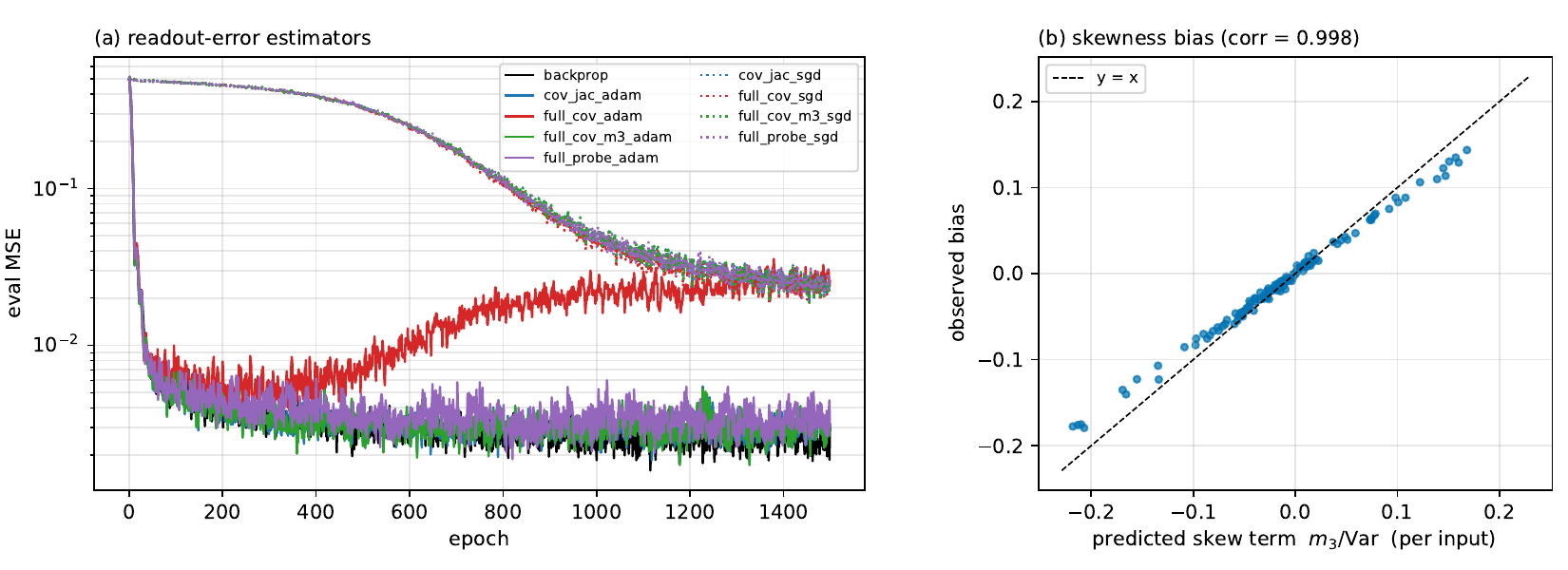}
 \caption{Readout-error estimation from forward covariances and verification of the skewness bias. (a) Evaluation MSE for \texttt{cov\_jac} and three \texttt{cov\_jac\_full} readout-error estimators, namely uncorrected covariance regression (\texttt{full\_cov}), third central moment correction (\texttt{full\_cov\_m3}), and the symmetric probe (\texttt{full\_probe}). Results using Adam and SGD are compared with backpropagation. (b) Observed readout-error bias versus the predicted skewness term $m_3 / \mathrm{Var}(y)$ for individual inputs. The dashed line denotes equality, and the Pearson correlation coefficient is 0.998.}
\label{fig:readout_error}
\end{figure}

I evaluate the predicted skewness bias and its corrections by replacing the readout error estimation in \texttt{cov\_jac\_full} with the three variants from Section~\ref{subsec:cov_jac_full}, which are uncorrected regression, third order moment correction, and the symmetric probe. 
All other conditions match the main comparison. 
The final MSE values as mean $\pm$ std over seeds 0, 1, and 2 were $0.00057 \pm 0.00009$ for the third order moment correction, matching backprop at $0.00057 \pm 0.00010$. 
The symmetric probe yielded $0.00090 \pm 0.00011$, and the uncorrected variant was $0.0241 \pm 0.0007$. 
Until around epoch 200, the uncorrected curve converges to approximately 0.006 at the same rate as the third order moment correction, but it drifts to 0.024 after epoch 500 (Fig.~\ref{fig:readout_error}(a)). 
This behavior confirms the prediction in Section~\ref{subsec:cov_jac_full} that the skewness bias remains after convergence and that Adam continuously amplifies it relatively via gradient scale normalization.
Indeed, when using SGD, all three estimators stay at the minimal optimization loss described in Section~\ref{subsec:ablation}, showing no performance gap. 
I also verified the bias term directly. 
The observed bias matches the predicted $m_3/\mathrm{Var}$ with a correlation of 0.998 for each input. 
Near convergence, the bias RMS of 0.062 exceeds the true error signal RMS of 0.049 (Fig.~\ref{fig:readout_error}(b)). 
Thus, the skewness bias is real and matches the predicted magnitude, and the third order moment correction alone raises the readout covariance estimation to the backpropagation-level.

\subsection{Benchmarking}
\label{subsec:benchmark}

To confirm that the main results are not task specific, I compare the five methods by changing only the tasks while maintaining the identical protocol from Section~\ref{subsec:setup}, including network architecture, $T$, epoch count, and optimizer selection as outlined in Table~\ref{tab:methods}. 
The three tasks are Friedman \#1 regression with a five dimensional input and targets normalized from -1 to 1, and binary classification of two moons and concentric circles using MSE learning with targets of $\pm 1$. 
The two classification tasks are linearly inseparable and cannot be solved unless credit assignment to the hidden layers functions properly. 
The datasets are fixed, and seeds vary only the initial weights and the noise during training. 
Table~\ref{tab:bench} presents the final MSE as mean $\pm$ std over seeds 0, 1, and 2, with classification accuracies in parentheses.

\begin{table}[tb]
\centering
\caption{Final performance on three benchmark tasks. Entries report the mean $\pm$ standard deviation of the final MSE over three different seeds for Friedman \#1 regression, Two Moons classification, and Concentric Circles classification. Friedman \#1 targets were normalized to $[-1,1]$, and the classification tasks used labels of $\pm 1$. Classification accuracies are shown in parentheses.}
\label{tab:bench}
\begin{tabular}{llll}
\hline
Method & Friedman \#1 & Two Moons & Concentric Circles \\
\hline
backprop & $0.00030 \pm 0.00001$ & $0.00035 \pm 0.00005$ (1.000) & $0.00202 \pm 0.00053$ (1.000) \\
\texttt{cov\_only} & $0.12043 \pm 0.00920$ & $0.80647 \pm 0.02375$ (0.854) & $0.71999 \pm 0.03417$ (0.943) \\
\texttt{cov\_deriv} & $0.07735 \pm 0.01048$ & $0.13931 \pm 0.01394$ (0.971) & $0.21531 \pm 0.00663$ (1.000) \\
\texttt{cov\_jac} & $0.00033 \pm 0.00006$ & $0.00044 \pm 0.00006$ (1.000) & $0.00242 \pm 0.00068$ (1.000) \\
\texttt{cov\_jac\_full} & $0.00033 \pm 0.00006$ & $0.00040 \pm 0.00015$ (1.000) & $0.00267 \pm 0.00041$ (1.000) \\
\hline
\end{tabular}
\end{table}

As shown in Table~\ref{tab:bench}, the final MSE of \texttt{cov\_jac} and \texttt{cov\_jac\_full} matches backprop within the range of variability between seeds across all three tasks, and all classification accuracies reach 1.000. 
The performance ranking among the methods is identical to Section~\ref{subsec:main_results}, where \texttt{cov\_deriv} achieves intermediate results and \texttt{cov\_only} ranks lowest. 
On the two moons task, \texttt{cov\_only} even fails classification with an accuracy of 0.854. Therefore, the conclusions in Section~\ref{subsec:main_results} are not unique to $\sin(x)$.

\section{Extended Evaluation}
\label{sec:extended}

In this section, the protocol from Section~\ref{subsec:setup} is kept unless stated otherwise, and the final errors and accuracies are reported as the mean $\pm$ standard deviation over three seeds.

\subsection{Sensitivity to the Ensemble Size, Noise Intensity, and Estimator Hyperparameters}
\label{subsec:sensitivity}

\begin{figure}[tb]
 \centering
    \includegraphics[width=1.0\textwidth]{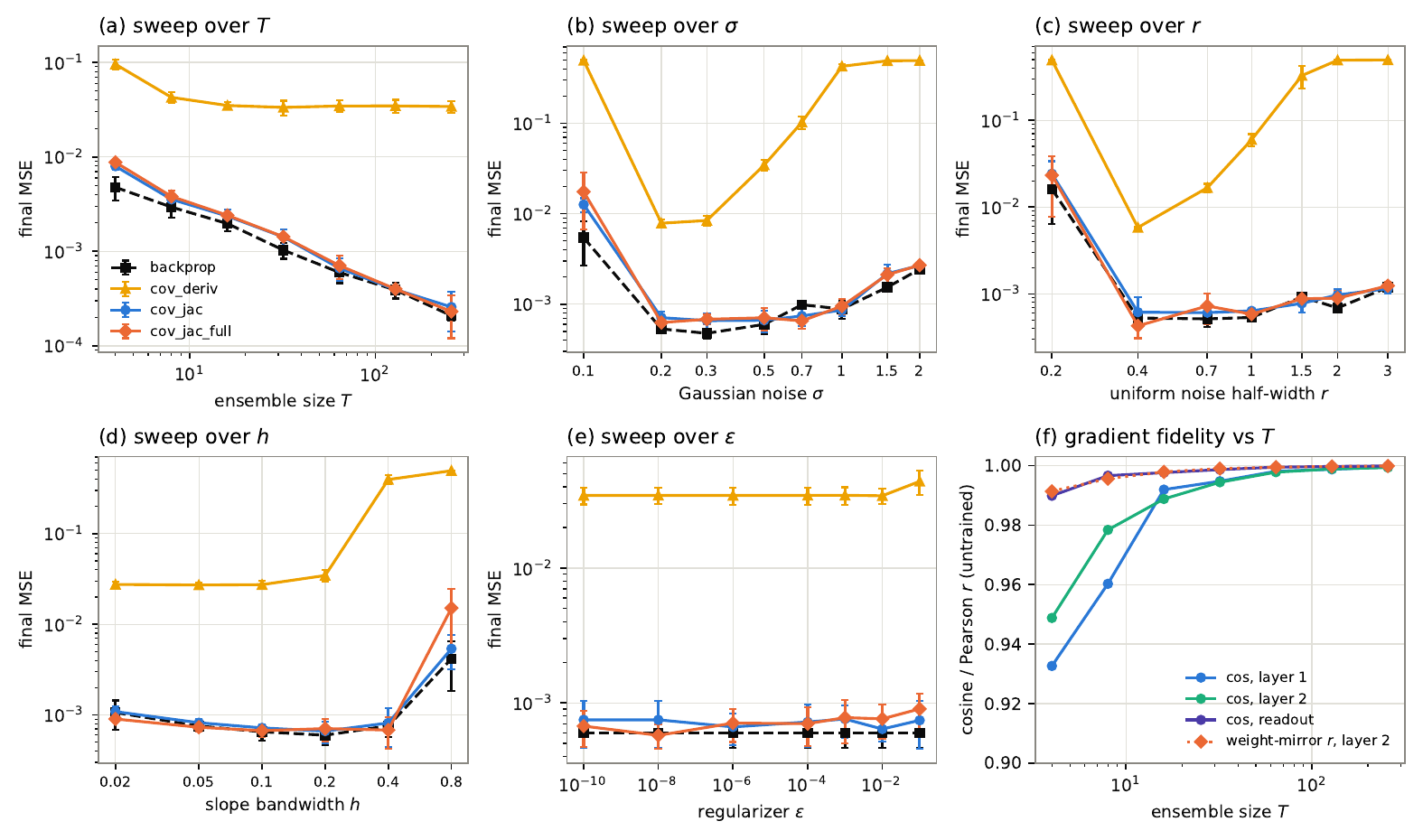}
 \caption{Sensitivity analysis on the $\sin(x)$ task. Panels (a)--(e) show the final evaluation MSE of backprop, \texttt{cov\_deriv}, \texttt{cov\_jac}, and \texttt{cov\_jac\_full} on a logarithmic scale over three seeds. Each panel varies one parameter while keeping the others at the values given in Section~\ref{subsec:setup}. The varied parameters are (a) the ensemble size $T$, (b) the Gaussian noise standard deviation $\sigma$, (c) the uniform noise half-width $r$, (d) the slope bandwidth $h$, and (e) the regularizer $\epsilon$. Panel (f) shows the cosine similarity between the \texttt{cov\_jac} update and the backpropagation reference update for each layer, together with the Pearson correlation between the estimated and true hidden-layer weights. These quantities are measured in the untrained network as $T$ varies.}
\label{fig:rev_sweeps}
\end{figure}

I swept $T$, $\sigma$, $r$, $h$, and $\epsilon$ one at a time around the setting of Section~\ref{subsec:setup} (Fig.~\ref{fig:rev_sweeps} (a)--(e)).
For each setting, I also measured the gradient fidelity of \texttt{cov\_jac} in the untrained network, as described in Section~\ref{subsec:gradient_fidelity} (Fig.~\ref{fig:rev_sweeps} (f)).
The final MSE of \texttt{cov\_jac} and \texttt{cov\_jac\_full} follows that of backprop from $T=4$ to $T=256$, with all three improving as $T$ grows.
The cost of learning statistics grows linearly with $T$ (Table~\ref{tab:methods}), so the required ensemble accuracy determines the sampling cost in these experiments.
The cosine similarity of the first hidden layer is $0.93$ at $T = 4$ and exceeds $0.99$ for $T \geq 16$, and the weight-mirror correlation remains above $0.99$ at every $T$.
At every $T$, the number of epochs that \texttt{cov\_jac} and backprop need to reach twice the final MSE of backprop differs by no more than 25\%.

For the noise intensity, the three methods match for $\sigma$ between $0.2$ and $1.0$, and for $r$ between $0.4$ and $3.0$.
Outside these ranges, all methods degrade together, including backprop.
At the low end, the noise becomes narrower than the slope window $h$, and at the high end, the response bumps broaden.
The final MSE of backprop, \texttt{cov\_jac}, and \texttt{cov\_jac\_full} changed little for $h$ from $0.02$ to $0.4$.
At $h = 0.8$, every method degraded, including the backprop reference, because its automatic differentiation uses the same estimator.
The regularizer $\epsilon$, which is added to the sample variance in the denominators of the credit and the weight mirror, had no measurable effect on the final MSE of \texttt{cov\_jac} and \texttt{cov\_jac\_full} from $10^{-10}$ to $10^{-1}$.
The weight-mirror denominator is pooled over the $N$ inputs, so $\epsilon$ is negligible there.
The $h$ and $\epsilon$ sweeps gave update cosine similarities above $0.98$ between \texttt{cov\_jac} and the backpropagation reference, except for $0.95$ in the first hidden layer at $h=0.02$. Weight-mirror correlations remained above $0.999$.
\texttt{cov\_deriv}, in contrast, saturated at an MSE of about $0.035$ for every $T \geq 16$ and worked only within a narrow window of noise intensity and $h$.
This indicates that its residual error is a bias rather than a finite-sample effect.

\subsection{Effect of Intra-layer Correlation}
\label{subsec:correlation}

The weight mirror of Section~\ref{subsec:cov_jac} recovers the forward weights when the fluctuations of different units in a layer are uncorrelated within an input.
If they are correlated, the univariate regression gives $\widehat W = W R$ with $R_{ki} = \mathrm{Cov}(z_k, z_i)/\mathrm{Var}(z_i)$, and the multivariate regression $\mathrm{Cov}(d, z)\,\mathrm{Cov}(z, z)^{-1}$ removes this bias at a cost of $O(H^3)$.
Both statements hold in the population limit with the regularizer $\epsilon$ neglected.
To quantify the effect, I replaced a fraction $\rho$ of the noise variance of each unit with a Gaussian component shared by all units of the layer within each sample, $\eta_k = \sigma(\sqrt{1-\rho}\,\varepsilon_k + \sqrt{\rho}\,\varepsilon_{\mathrm{c}})$, where $\rho = 0$ corresponds to the independent noise used elsewhere in this paper.
The effect grows monotonically with $\rho$.
At $\rho = 0.3$, the measured correlations between the fluctuations of the units within the first and second hidden layers are $0.01$ and $0.04$, the correlation of the hidden-layer weight mirror is still $0.97$, and the final MSE of \texttt{cov\_jac} is unchanged.
At $\rho = 0.9$, these fluctuation correlations are $0.14$ and $0.46$, the weight-mirror correlation falls to $0.50$, and the cosine similarity of the first hidden layer decreases from $0.998$ to $0.33$. The final MSE of the univariate \texttt{cov\_jac} rises to $0.023$ against $0.0013$ for backprop, whereas the final MSE with the multivariate weight mirror remains at $0.0018$.
In the networks of Sections~\ref{sec:experiment} and \ref{subsec:sensitivity}, the measured fluctuation correlation is below $0.005$ because the intrinsic crossing noise of each unit dominates its fluctuations.
Section~\ref{subsec:deep} examines the case in which such correlation arises on its own in deeper networks.

\subsection{Comparison with Alternative Forward-only Rules and a Conventional Weight Mirror}
\label{subsec:baselines}

\begin{table}[tb]
\centering
\caption{Comparison with alternative forward-only rules on the same NNN, data, and forward samples. The first four columns report final MSE for $\sin(x)$ and the three benchmarks in Table~\ref{tab:bench}, with classification accuracy in parentheses when below $1.0$. The last column reports the range of cosine similarities between hidden-layer updates and the backpropagation reference update across the four hidden layers of an MNIST network with the architecture of Section~\ref{subsec:deep}, pretrained for 300 epochs by backpropagation on 250 images. This measure applies to \texttt{cov\_jac}, FA, and DFA, which propagate layer-wise errors. PEPITA failed to learn $\sin(x)$ and was not run on the other benchmarks.}

\label{tab:rev_baselines}
\footnotesize
\setlength{\tabcolsep}{4pt}
\resizebox{\textwidth}{!}{%
\begin{tabular}{lccccc}
\hline
 & \multicolumn{4}{c}{final MSE} & hidden-layer cosine \\
Method & $\sin(x)$ & Friedman \#1 & Two Moons & Circles & (MNIST, four hidden layers) \\
\hline
backprop & $0.00057 \pm 0.00010$ & $0.00030 \pm 0.00001$ & $0.00035 \pm 0.00005$ & $0.00202 \pm 0.00053$ & 1 \\
\texttt{cov\_jac} & $0.00056 \pm 0.00006$ & $0.00033 \pm 0.00006$ & $0.00044 \pm 0.00006$ & $0.00242 \pm 0.00068$ & 0.80 to 0.94 \\
FA & $0.00084 \pm 0.00017$ & $0.00042 \pm 0.00004$ & $0.00113 \pm 0.00012$ & $0.00637 \pm 0.00053$ & $-0.07$ to $0.00$ \\
DFA & $0.00081 \pm 0.00009$ & $0.00037 \pm 0.00005$ & $0.00093 \pm 0.00025$ & $0.00518 \pm 0.00141$ & $-0.02$ to $0.08$ \\
node perturbation & $0.0266 \pm 0.0034$ & $0.143 \pm 0.007$ & $0.184 \pm 0.005$ (0.956) & $0.288 \pm 0.017$ (0.997) & -- \\
PEPITA & $0.459 \pm 0.002$ & -- & -- & -- & -- \\
\hline
\end{tabular}}
\end{table}

Table~\ref{tab:rev_baselines} compares the main alternatives of Section~\ref{subsec:2.3}, implemented on the same NNN with the same data and forward samples.
Feedback alignment (FA) replaces $\widehat W$ in the recursion of Section~\ref{subsec:cov_jac} with a fixed random matrix, and direct feedback alignment (DFA) projects the output error to each hidden layer through fixed random matrices.
Node perturbation injects an external perturbation of amplitude $0.1$ into every unit and uses a common-random-number estimator.
PEPITA uses an error-modulated second forward pass, and I report the best of three modulation factors.
FA and DFA use Adam, and node perturbation uses Adam on $\sin(x)$ and SGD on the benchmarks.
PEPITA uses SGD.
FA and DFA reach a final MSE of $1.1$ to $2.6$ times that of \texttt{cov\_jac} and also solve the two classification tasks.
The final MSE of node perturbation is 50 to 430 times that of \texttt{cov\_jac}, and PEPITA did not learn in the settings I explored.

The rightmost column of Table~\ref{tab:rev_baselines} compares the update directions in a four-hidden-layer network with the architecture of Section~\ref{subsec:deep}, pretrained for 300 epochs by backpropagation on a class-balanced subset of 250 MNIST images. The cosine similarity to the backpropagation reference update is $0.80$ to $0.94$ for \texttt{cov\_jac} and below $0.1$ in absolute value for FA and DFA across the hidden layers. 
The same ranking holds for the untrained network.
FA and DFA instead rely on the gradual alignment of the forward weights with the fixed feedback matrices during their training \cite{Lillicrap2016random}, which was not measured here.

For the $\sin(x)$ task, I constructed a dedicated-phase weight-mirror baseline modeled on \cite{akrout2019deep}.
It retains the recursion, exponential moving average, and Kolen-Pollack tracking of \texttt{cov\_jac}, changing only the weight-mirror measurement.
In the dedicated phase, $K$ isotropic Gaussian perturbation vectors are injected into each layer's input, and the resulting pre-activations are regressed on these perturbations.
The mean final MSE over three seeds ranges from $0.00066$ to $0.00097$ across variants.
The intrinsic weight mirror, estimated from the $T = 64$ inference samples of each input, maintains a correlation above $0.996$ with the true hidden-layer weights throughout training without additional forward evaluations.
With the dedicated phase, the minimum of this correlation during training is $0.90$ for $K = 256$ and $0.41$ for $K = 16$. Using $K = 256$ costs $512$ additional linear-layer evaluations per update and requires an injection port at every layer input.
The advantage of the intrinsic fluctuations is thus the removal of the dedicated phase and its sampling cost, rather than a higher final accuracy.

\subsection{Deeper Networks, Multiclass Classification, and the Readout Estimator}
\label{subsec:deep}

\begin{table}[tb]
\centering
\caption{MNIST classification with the cross-entropy loss (two to four hidden layers of 256 units, three seeds, settings in the text). (a) Depth dependence. Final test accuracy and the minimum, over the hidden layers, of the cosine similarity between the hidden-layer update and the backpropagation reference update at the end of training, averaged over the seeds. (b) Readout-error estimators with two hidden layers, with the hidden-layer credit of \texttt{cov\_jac} in every row. Final test accuracy and the cosine similarity between the estimated and the analytical readout error at the end of training. The runs of (b) are separate from those of (a) and use their own seeds.}
\label{tab:rev_deep}
\footnotesize
\setlength{\tabcolsep}{4pt}
\textbf{(a) Depth dependence}\par\smallskip
\begin{tabular}{lccccc}
\hline
 & backprop & \multicolumn{2}{c}{\texttt{cov\_jac}} & \multicolumn{2}{c}{\texttt{cov\_jac} + noise field} \\
hidden layers & accuracy & accuracy & cosine & accuracy & cosine \\
\hline
2 & $0.891 \pm 0.002$ & $0.887 \pm 0.004$ & 0.80 & $0.889 \pm 0.001$ & 0.88 \\
3 & $0.886 \pm 0.003$ & $0.880 \pm 0.002$ & 0.58 & $0.887 \pm 0.003$ & 0.81 \\
4 & $0.882 \pm 0.003$ & $0.863 \pm 0.007$ & 0.38 & $0.871 \pm 0.002$ & 0.83 \\
\hline
\end{tabular}
\par\medskip
\textbf{(b) Readout-error estimators with two hidden layers}\par\smallskip
\begin{tabular}{lcc}
\hline
Readout error & test accuracy & cosine to the analytical error \\
\hline
analytical gradient $\mathrm{softmax}(\bar{\mathbf{y}}) - \mathbf{t}$ (\texttt{cov\_jac}) & $0.888 \pm 0.002$ & 1.000 \\
Eq.~(\ref{eq:readout_reg}), uncorrected & $0.621 \pm 0.043$ & 0.14 \\
Eq.~(\ref{eq:readout_reg}) with correction (\texttt{cov\_jac\_full}) & $0.203 \pm 0.112$ & 0.09 \\
Eq.~(\ref{eq:mv_readout}), multivariate, uncorrected & $0.889 \pm 0.002$ & 0.80 \\
symmetric probe, Gaussian, $\alpha = 0.2$ & $0.891 \pm 0.003$ & 0.29 \\
\hline
\end{tabular}
\end{table}

I trained NNNs with 784 inputs, two to four hidden layers of 256 units, and 10 outputs on a class-balanced subset of MNIST using the cross-entropy loss.
The settings are $T = 64$, Gaussian noise with $\sigma = 0.5$, Adam with a learning rate of $10^{-3}$, full-batch updates, and 400 epochs.
The data are a class-balanced subset of 2000 images drawn once with a fixed seed from the original MNIST training set \cite{lecun1998gradient} and split by a fixed permutation into 1000 training and 1000 test images.
Pixel values are scaled to $[-1, 1]$, and the same images are used for every learning seed.
Table~\ref{tab:rev_deep}(a) shows that \texttt{cov\_jac} is within 0.6 percentage points of backprop with two or three hidden layers.
With four hidden layers, the gap increases to 1.9 points, and the spread across seeds becomes larger.
The weight mirrors estimated from the second and deeper hidden layers degrade with depth.
Their minimum correlation at the end of training is $0.997$, $0.981$, and $0.943$ for two, three, and four hidden layers.
The minimum hidden-layer cosine similarity also falls from $0.80$ to $0.38$ as depth increases from two to four hidden layers.
Shared upstream fluctuations induce the intra-layer correlation described in Section~\ref{subsec:correlation} even without an injected common component, and this effect grows with depth.

The noise field of Section~\ref{subsec:network} specifies only the marginal distribution of the noise of each unit.
The dependence between samples can be designed to reduce the effect of intra-layer correlations.
While the weight mirror of layer $l$ is measured, the noise of every upstream layer is tiled along the sample axis with a period $P = 16$, $\eta_{(m)} = \eta_{(m \bmod P)}$. The $T = 64$ samples of an input then contain $P$ distinct upstream noise configurations, each repeated $T/P = 4$ times.
Within each phase class of samples that share an upstream configuration, the drive to layer $l$ is fixed, and its units fluctuate independently.
The covariance and variance of Section~\ref{subsec:cov_jac} are therefore accumulated after centering within each phase class, which removes the correlation bias in the population limit.
One weight mirror is measured per epoch in rotation over the hidden layers, and the remaining weight mirrors are updated by the Kolen-Pollack method in that epoch.
The cost is the smaller number of distinct upstream configurations in the measuring epochs and the centering within phase classes, which requires $P$ running means per unit instead of one.
With this noise field, these weight mirrors keep a correlation of $1.000$ (to three decimals) at every depth. Table~\ref{tab:rev_deep}(a) shows that the minimum hidden-layer cosine similarity stays above $0.8$ at every depth.
At four hidden layers, the spread of the accuracy across seeds shrinks, but a gap of about one percentage point to backprop remains.

Table~\ref{tab:rev_deep}(b) tests the readout-error estimators of Section~\ref{subsec:cov_jac_full} under the cross-entropy loss with two hidden layers and the hidden-layer credit of \texttt{cov\_jac} in every case.
The per-output univariate regression fails with and without the third-moment correction. The ten logits fluctuate jointly through the shared hidden samples, and the scalar third-moment correction is incorrect for the cross-entropy.
The multivariate regression of Eq.~(\ref{eq:mv_readout}), with a ridge term of $10^{-3}$ times the mean readout variance and without any moment correction, and the symmetric probe both achieve test accuracy comparable to that obtained with the analytical readout error, although their cosine similarity to the analytical error decreases during training.
The ridge term regularizes the estimator, whereas the exact population relation of Section~\ref{subsec:cov_jac_full} is stated for the unregularized case.
The third-moment-corrected \texttt{cov\_jac\_full} should be used only with a quadratic loss and a single readout unit, as in Section~\ref{sec:experiment}.

\section{Hardware-oriented Analysis}
\label{sec:hardware_implementation}

\subsection{Resource Requirements}
\label{subsec:resource_requirements}

Table~\ref{tab:resource} compares the forward learning rules of Section~\ref{sec:forward_only_learning_rule} with backpropagation by counts of multiply-accumulate (MAC) operations and memory words.
The table presumes a digital, event-driven implementation in which the weights reside in ordinary single-port memory, as in an FPGA, where a transposed read requires either a second copy of the weights or a transposed access pattern, and backpropagation needs a separate backward phase.
The learning statistics are the running sums of Section~\ref{subsec:cov_deriv} and the covariance $\mathrm{Cov}(d, z)$ of the weight mirror, and estimating the readout error from the covariance as in Section~\ref{subsec:cov_jac_full} adds one accumulation per readout unit, $\overline{y^3}$.
No transposed access to the forward weights and no dedicated perturbation phase are needed, but credit is still propagated recursively through the estimated weight mirrors, once per input, and this recursion is counted.
The counts for backprop on the NNN follow the sample-wise backward pass of automatic differentiation, which stores the per-sample activations and propagates the error for every sample.
Because the backpropagation reference uses the sample-averaged slope of Section~\ref{subsec:crossing}, its error signal is the same for all samples of an input.
An implementation that exploits this feature would reduce its full-precision MACs and its buffer.
The counts are therefore specific to the implementations compared and are not lower bounds.

\begin{table}[tb]
\centering
\caption{Operation and memory counts per update for two hidden layers of width $H=64$, $N=128$ inputs, and $T=64$ samples per input. Counts include forward sampling, random-number generation, readout averaging, learning statistics, and optimizer state, but exclude optimizer arithmetic. The deterministic baseline uses the same network size with $T=1$ and Adam. NNN methods use the optimizers listed in Table~\ref{tab:methods}. Binary-gated MACs use AND-gated additions, whereas full-precision MACs require multipliers. Each crossing reuses the previous noise sample, requiring one new $b$-bit draw per hidden unit and sample. The generator state counts one 16-bit LFSR per hidden unit. Divisions are for estimator normalization. Memory counts exclude the forward weights and biases. The NNN backpropagation buffer stores activations and errors for all $NT$ samples for the sample-wise backward pass. These are operation and memory counts, not measurements of area, power, or latency.
}
\label{tab:resource}
\footnotesize
\setlength{\tabcolsep}{4pt}
\begin{tabular}{lcccc}
\hline
 & deterministic NN & NNN & NNN & NNN \\
 & backprop, Adam & backprop, Adam & \texttt{cov\_deriv}, SGD & \texttt{cov\_jac}, Adam \\
\hline
Forward-pass full-precision MACs & 0.54M & 0.52M & 0.52M & 0.52M \\
Forward-pass binary-gated MACs & 0 & 34.1M & 34.1M & 34.1M \\
Random-number draws & 0 & 1.05M & 1.05M & 1.05M \\
Learning full-precision MACs & 1.07M & 34.1M & 0 & 0.53M \\
Learning binary-gated MACs & 0 & 34.6M & 37.2M & 70.3M \\
Divisions (estimators) & 0 & 0 & 16k & 4.2k \\
\hline
Total full-precision MACs & 1.61M & 34.6M & 0.52M & 1.06M \\
Total binary-gated MACs & 0 & 68.7M & 71.3M & 104M \\
\hline
Memory for learning statistics (words) & 20.6k & 1.05M & 0.5k & 8.4k \\
Memory for optimizer state (words) & 8.4k & 8.4k & 0 & 8.4k \\
Memory for generator state (words) & 0 & 128 & 128 & 128 \\
Total additional memory (words) & 29.1k & 1.06M & 0.6k & 17k \\
\hline
Transposed weight path & required & required & not required & not required \\
\hline
\end{tabular}
\end{table}

Within the NNN, the forward rules avoid the transposed weight path and the per-sample buffer.
\texttt{cov\_jac} needs 0.53M full-precision MACs for the credit recursion, which occurs once per input regardless of $T$, and 17k words of additional memory, against 34.1M full-precision MACs on the transposed weight path and a per-sample buffer of 1.05M words for backprop.
Relative to the deterministic network, the forward sampling dominates, because \texttt{cov\_jac} performs a similar number of full-precision multiplications but about $10^{8}$ binary-gated accumulations per update, which is the price of the $T = 64$ samples.
The forward rules therefore offer a different trade-off rather than a demonstrated saving.
Almost all of their operations are AND-gated accumulations, comparisons, and counters, so a learning core can be built largely from ordinary digital logic with few multiplier blocks and a small memory, at the price of more samples per update.
Note that this does not imply lower energy consumption or latency than backpropagation on a deterministic network.

The comparison also depends on the hardware architecture.
In a resistive crossbar or another in-memory design, the same synaptic array can be read in both directions, so the transpose is available at little extra cost.
In such architectures, the forward rules still avoid the sample-wise backward pass and its per-sample buffer, but they save nothing on the weight path, and the $T$-fold forward sampling remains as a cost.

\subsection{Noise Selection for Sparseness}
\label{subsec:noise_selection}

The computational and memory requirements listed in Section~\ref{subsec:resource_requirements} do not depend on the noise distribution. 
Furthermore, in this configuration, where the local slope is estimated by the kernel density estimation shown in Section~\ref{subsec:crossing}, the analytical form of the noise distribution appears nowhere in inference or learning.
That is, changing the noise distribution does not alter the network operations. 
I confirm this through a replication experiment using the identical protocol but replacing the noise with a uniform distribution $p(\xi) = \frac{1}{2r}\,\mathbf{1}\{\lvert\xi - c\rvert \le r\}$, where $c$ is the center and $r$ is the half width, with $c = 0$ and $r = 1.0$ in the experiment. 
As shown in Table~\ref{tab:uniform}, the final MSE of \texttt{cov\_jac} and \texttt{cov\_jac\_full} matches backprop even with uniform noise, and the performance ranking among methods remains identical to Section~\ref{subsec:main_results}. 
The agreement between the estimated slope and the analytical $\bar\phi'$ was also reproduced, yielding 0.054 compared to 0.058. 
Therefore, the noise source can be selected based on implementation cost rather than learning performance.

\begin{table}[tb]
\centering
\caption{Effect of the noise distribution on final regression performance. Values report the mean $\pm$ standard deviation of the final MSE over three different seeds under Gaussian noise with $\sigma=0.5$ and uniform noise centered at $c=0$ with half-width $r=1.0$. The Gaussian-noise results are reproduced from Table~\ref{tab:performance}, and all other experimental conditions are unchanged.}
\label{tab:uniform}
\begin{tabular}{lll}
\hline
Method & Gaussian ($\sigma=0.5$, Table 2 reproduced) & Uniform ($r=1.0$) \\
\hline
backprop & $0.00057 \pm 0.00010$ & $0.00050 \pm 0.00002$ \\
\texttt{cov\_only} & $0.09583 \pm 0.00724$ & $0.12688 \pm 0.00534$ \\
\texttt{cov\_deriv} & $0.03464 \pm 0.00558$ & $0.05765 \pm 0.00951$ \\
\texttt{cov\_jac} & $0.00056 \pm 0.00006$ & $0.00046 \pm 0.00001$ \\
\texttt{cov\_jac\_full} & $0.00057 \pm 0.00009$ & $0.00048 \pm 0.00005$ \\
\hline
\end{tabular}
\end{table}

From this perspective, there are two essential reasons to choose uniform noise in digital hardware implementations. 
First, uniform random numbers can be generated using only an LFSR via shifts and XOR operations.
This removes the extra circuits for the Box-Muller or CLT approximations needed for Gaussian noise. 
Second, the statistically equivalent activation function has compact support. 
With uniform noise, $F(d) = (d - c + r)/(2r)$ holds within the range $\lvert d - c \rvert < r$. 
From the general formula $\bar\phi = 2F(1-F)$ in Section~\ref{subsec:crossing}, the expected response reduces to a strict parabola, and the local derivative simplifies to a linear expression.
\begin{eqnarray}
\bar\phi(d) &=& \frac{1}{2}\left[1 - \left(\frac{d - c}{r}\right)^{2}\right]_{+} \\
\bar\phi'(d) &=& -\frac{d - c}{r^{2}} \quad (\lvert d - c \rvert < r, \text{ and } 0 \text{ for } \lvert d - c \rvert > r).
\end{eqnarray}
At $\lvert d - c \rvert = r$, the derivative is undefined.
The expected response is strictly 0 when $\lvert d - c \rvert \geq r$, meaning that units far from the threshold maintain exactly zero spikes. 
With Gaussian noise, the firing probability is positive everywhere, meaning this sparsity cannot be achieved. 
Compact support directly leads to activity and power sparsity in event driven implementations.
The simplicity of the analytical expressions is a secondary benefit because they are not used in this configuration employing the estimated slope.

\begin{figure}[tb]
 \centering
    \includegraphics[width=0.9\textwidth]{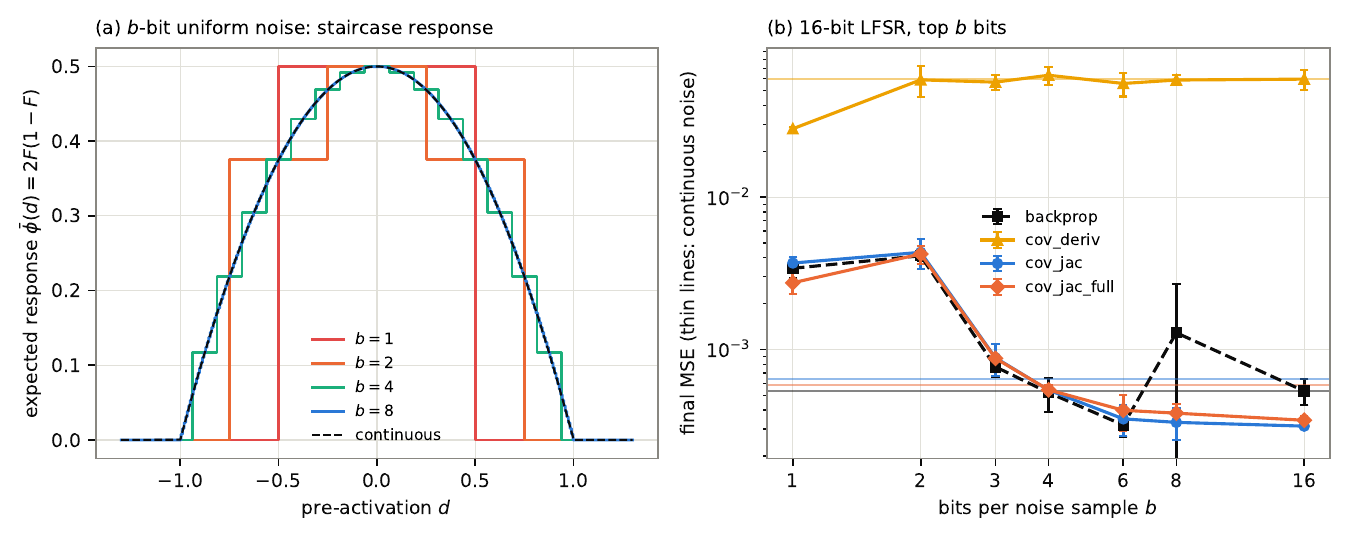}
 \caption{Learning under $b$-bit digitally generated uniform noise ($r = 1.0$, with all other settings as in Table~\ref{tab:uniform}). (a) Expected response $\bar\phi = 2F(1-F)$ for independent $b$-bit quantized uniform noise. The response is a staircase approximation of the continuous parabola (dashed). (b) Final MSE (mean $\pm$ standard deviation over three seeds) versus $b$ for the top $b$ bits of a maximal-length 16-bit LFSR clocked $b$ times per sample. Thin horizontal lines mark the results of the same experiment with continuous uniform noise.}
\label{fig:rev_discrete}
\end{figure}

A digital random-number generator produces $b$-bit integers rather than the continuous uniform distribution assumed above, so the noise has a discrete distribution.
Figure~\ref{fig:rev_discrete}(a) shows the staircase response for independent draws from this quantized uniform distribution.
To determine the required noise resolution, the uniform-noise experiment of Table~\ref{tab:uniform} was repeated using a maximal-length 16-bit LFSR. 
Each sample used its top $b$ bits, with the LFSR advanced by $b$ clock cycles between samples. 
Independent $b$-bit quantized samples served as a control to distinguish quantization effects from serial correlation in the LFSR sequence.
Figure~\ref{fig:rev_discrete}(b) shows that for backprop, \texttt{cov\_jac}, and \texttt{cov\_jac\_full}, one or two bits increase the final MSE five- to eightfold, three bits increase it by about half, and with four or more bits the errors are no larger than in the continuous case, apart from one backprop seed at eight bits.
\texttt{cov\_deriv} is not degraded, because its error is dominated by its own systematic bias.
The similar degradation of backprop and the weight-mirror correlation above $0.995$ at every bit width suggest that the reduced resolution of the staircase response, rather than a failure of the covariance estimators, is the main cause.
The independent quantized samples behave in the same way, and the LFSR is better at six bits and above.

Finally, noise is a design variable not only in its distribution but also in its spatial assignment. 
The noise field described in Section~\ref{subsec:network} implies that the noise assignment determines which units receive credit, providing design flexibility to suppress intra-layer correlations and directly couple update gating with power consumption gating. 
Section~\ref{subsec:deep} uses this flexibility along the sample axis to suppress intra-layer correlations, and the coupling with power gating is left for future work.

\section{Discussion}
\label{sec:discussion}

The proposed learning rules use forward statistics for credit assignment without weight transport or a dedicated perturbation phase. 
On the regression and binary classification tasks with squared loss, \texttt{cov\_jac} and the third-moment-corrected \texttt{cov\_jac\_full} achieve accuracy comparable to backpropagation. 
On the MNIST subset, \texttt{cov\_jac} with the analytical readout error remains within two percentage points of backpropagation at all tested depths.

The evaluation is limited to networks with up to four hidden layers on a subset of MNIST. 
At four hidden layers, a gap of about one percentage point to backpropagation remains even with the noise field, and intra-layer correlations increase with depth (Section~\ref{subsec:deep}). 
Further experiments are needed to determine whether the sample-axis structure of the noise field can maintain accurate weight mirrors in deeper and wider networks and on larger datasets.

The hardware benefits also remain to be tested. 
Section~\ref{sec:hardware_implementation} estimates the computation and memory required for learning without transpose access to the forward weights. 
These estimates depend on the implementations compared, and stochastic sampling adds a cost relative to a deterministic network. 
Hardware measurements are needed to assess the resulting area, power consumption, and latency.

Finally, the proposed approach is neuromorphic in one functional sense. 
It uses neural variability as a computational resource rather than treating it as a disturbance, as discussed in Section~\ref{sec:intro}. 
The same fluctuations used for inference also provide learning signals, without exact weight transport or a separate perturbation phase. 
This biological inspiration does not extend to the learning mechanism itself. Covariance normalization, a stored weight mirror, recursive error propagation, and an adaptive optimizer are not intended to model biological circuitry.

\section{Conclusion}
\label{sec:conclusion}

This paper demonstrates that the structure of backpropagation can be reconstructed in a noise modulated neural network (NNN) solely from the covariance statistics of the forward fluctuations. 
The proposed \texttt{cov\_jac} rule estimates weights from forward perturbations in crossing activations and recurses them layer wise. 
Furthermore, \texttt{cov\_jac\_full} estimates the readout error from the covariance, generalizing \texttt{cov\_jac} to a configuration that completely eliminates analytical loss derivatives. 
Both rules approximate exact gradients without transporting or transposing weights, achieving accuracy comparable to backpropagation in regression and binary classification tasks with the squared loss. 
\texttt{cov\_jac} with the analytical readout error further scales to a class-balanced subset of MNIST with up to four hidden layers, where the noise field of Section~\ref{subsec:deep} reduces the gap to backpropagation to about one percentage point. On the other hand, the third-central-moment correction of \texttt{cov\_jac\_full} is limited to quadratic losses with a single readout unit.
These learning rules naturally integrate existing ideas that avoid weight transport into NNN dynamics without a dedicated perturbation phase, using the noise that the network already needs for inference. 
Additionally, because the required operations are confined to a few primitives centered around accumulators and comparators, the overall configuration maps onto standard digital elements, although I claim architectural simplicity rather than a measured advantage in power or latency.
In conclusion, this paper shows that the NNN is a mathematical model that combines a functional use of neural noise with engineering applicability in neuromorphic computing.

\section*{Acknowledgement and Funding}
This work was supported by the Japan Science and Technology Agency (JST) Fusion Oriented Research for Disruptive Science and Technology (FOREST) Program under Grant JPMJFR242H.
This work was supported by the Japan Society for Promotion of Science (JSPS) KAKENHI Grant Number 25H02619.
The author declares no competing interests.

\roles{
Shuhei Ikemoto \orcid{0000-0003-4885-8746}

Conceptualization, Methodology, Software, Validation, Formal analysis, Investigation, Visualization, Writing – Original Draft, Writing – Review \& Editing, Funding acquisition
}

\data{
The data that support the findings of this study are openly available at the following URL: https://github.com/ikesyu/nnn-covariance-credit-assignment. The repository contains the complete source code of the Noise-modulated Neural Network library together with the scripts that reproduce all tables and figures reported in this paper, including the extended evaluation of Section~\ref{sec:extended}, released under the MIT License.
}

\small
\bibliographystyle{unsrt}
\bibliography{reference_rev}

\end{document}